\documentclass[preprint,3p]{elsarticle}

\usepackage[colorlinks]{hyperref}
\usepackage[linesnumbered,ruled]{algorithm2e}
\usepackage{cleveref,amssymb, graphicx, threeparttable, booktabs, multirow, amsthm, amsmath,epic,arydshln, makecell,colortbl,subfigure}
\makeatletter
\AtBeginDocument{\Hy@breaklinkstrue}
\makeatother

\journal{\phantom{a}}

\newtheorem{prop}{Proposition}[section]
\newtheorem{corollary}{Corollary}[section]
\newdefinition{define}{Definition}[section]

\newcommand{\s}{\mathbf}
\newcommand{\w}{\widetilde}

\newcommand{\ds}{\displaystyle}
\newcommand{\st}{\text{s.t.}}

\definecolor{gray1}{gray}{0.56}%
\definecolor{gray11}{gray}{0.62}
\definecolor{gray2}{gray}{0.68}%
\definecolor{gray21}{gray}{0.74}
\definecolor{gray3}{gray}{0.80}%
\definecolor{gray31}{gray}{0.86}
\definecolor{gray4}{gray}{0.92}%

\begin{document}

\begin{frontmatter}



\title{Feature selection for classification with class-separability strategy and data envelopment analysis}

 \author[a,b]{Yishi Zhang}
 \ead{zhang685@wisc.edu}
 \author[a]{Chao Yang}
 \ead{chao\_yang@hust.edu.cn}
  \author[a]{Anrong Yang}
  \ead{anrong\_yang@126.com}
 \author[a]{Chan Xiong}
 \ead{xiongchan0113@hust.edu.cn}
 \author[a,c]{Xingchi Zhou}
 \ead{xczhou@uw.edu}
 \author[a]{Zigang Zhang\corref{cor1}}
 \ead{zigang\_zhang@hust.edu.cn}
 \address[a]{School of Management, Huazhong University of Science and Technology, Wuhan 430074, China}
 \address[b]{Wisconsin School of Business, University of Wisconsin-Madison, Madison, WI 53706, USA}
 \address[c]{Foster School of Business, University of Washington, Seattle, WA 98195, USA}
 \cortext[cor1]{Corresponding author at: School of Management, Huazhong University of Science and Technology, 1037 Luoyu Road, Hongshan District, Wuhan 430074, China. Tel.: +86 27 87556463.}

\begin{abstract}
In this paper, a novel feature selection method is presented, which is based on Class-Separability (CS) strategy and Data Envelopment Analysis (DEA). To better capture the relationship between features and the class, class labels are separated into individual variables and relevance and redundancy are explicitly handled on each class label. Super-efficiency DEA is employed to evaluate and rank features via their conditional dependence scores on all class labels, and the feature with maximum super-efficiency score is then added in the conditioning set for conditional dependence estimation in the next iteration, in such a way as to iteratively select features and get the final selected features. Eventually, experiments are conducted to evaluate the effectiveness of proposed method comparing with four state-of-the-art methods from the viewpoint of classification accuracy. Empirical results verify the feasibility and the superiority of proposed feature selection method. 
\end{abstract}


\end{frontmatter}


\section{Introduction}
\label{Intro}
The explosion of large datasets in many fields poses unprecedented challenges to pattern recognition and data mining. Not only is the scale of samples getting larger, but also new types of data become prevalent. For example, tremendous new computer and Internet applications generate large amounts of types of data at an exponential rate in the world. It is thus realized that feature selection is an indispensable component \cite{InfoSourceSel}.
Feature selection is a process of selecting a subset of original features according to certain criteria. It is an important and frequently used technique for dimension reduction. It reduces dimensionality by removing irrelevant, redundant, or noisy features, thus bringing
about significant effects for applications: reducing the cost of data acquisition and collection, speeding up a learning algorithm, improving
learning accuracy, and resulting in better model comprehensibility \cite{MI_DSS}.

 Roughly speaking, there are two types of feature selection methods: Filter and wrapper \cite{2,4}. \footnote{In literature, feature selection could also be categorized into three types: Embedded, wrapper, and filter methods. Here we take embedded method as a wrapper.} Wrapper selects features relying on performance estimated by a specific learning method, thus its computational complexity as well as the quality of the selected features is highly related to the learning methods. In addition, wrapper suffers from poor generalization to other learning methods. Filter selects features in terms of the classifier-agnostic criteria (e.g. Fisher score \cite{6}, $\chi ^2$-test \cite{2,8}, Mutual Information (MI) \cite{9,11,12}, symmetrical uncertainty \cite{13}, Hilbert-Schmidt operator \cite{Le_Song}), where MI is an effective and efficient information theoretic metric and thus one of the most widely used feature selection criterion, although it is shown that mutual information can not always guarantee to decrease the misclassification probability \cite{Limit_MI}. Thus, it may be more generalized and efficient than wrapper \cite{13,9,14,15}. In this paper, we only focus on filter methods.

Regarding the search process, feature selection can also be categorized into two types. One searches the subset by taking it as an combinatorial problem rather than only investigating the variety of the discriminative abilities from individual perspective, while the other constructs the feature subset by ranking features in terms of their discriminative abilities. An typical example of the former is Correlation-based Feature Selection method (CFS) \cite{CFS}, which only cares the quality of the feature subsets and will not discriminate features from an individual perspective. Since finding an optimal feature subset in feature selection has been shown to be NP-hard \cite{NPHard,NPComplete}, it is usually intractable to conduct exhaustive search even with medium-scale datasets. On the other hand, optimal feature subsets according to specific criteria of filters may not be so effective in promoting learning accuracy since the criteria is classifier-irrelevant, thus impairing the enthusiasm for finding effective branch and bound methods or approximation methods that can guarantee optimal solutions or near-optimal ones in theory. On the contrary, the latter makes a trade-off between the quality of features and the computational time by inducing the individual perspective \cite{MIFS,MIFS-U,CMIM,13}. Many very fast feature selection methods that belong to this kind greedily select features in a sequential manner, and perform most efficiently with very simple heuristics, such as intra- and inter-class distances \cite{Relief,Var_Of_relief}, Maximum Mutual Information \cite{MIM}, etc. However, such heuristics sort features only in terms of the magnitude of the relevance to the class (i.e. class-relevance), and performs inferiorly especially on datasets with high dimensionality \cite{DEAFS}. This turns out to be that they never consider feature redundancy and thus leading to ``the $m$ best features are not the best $m$ features'' \cite{mRMR30}. Since the existence and effect of feature redundancy were pointed out \cite{KollerSahami,Kohavi_John,9,11}, how to select more relevant and less redundant features has been a hot issue in literature \cite{CFS,mRMR2,mRMR30,13,CMIM,MIFS-U,11,Zhangis_2}.

A series of heuristics are proposed to make an explicit trade-off between high class-relevance and low redundancy by applying different measures and parameters \cite{MIFS,MIFS-U,mRMR2,mRMR4,mRMR14,mRMR30,13,cluster_based}. A typical example is feature selection methods with minimal Redundancy and Maximal Relevance (mRMR) criterion \cite{MIFS,mRMR2,9,MIFS-U} which explicitly applies the relevance measure $D=I(F,C)$ and the redundancy measure $R=\beta\cdot\sum_{F,F_s\in \s{S}}I(F,F_s)$ (where $I$ denotes MI), and uses $J=\max(D-R)$ or $J=\max(D/R)$ to evaluate and select features from the individual perspective in a greedy manner. Beside this, there are several representative feature selection methods making an implicit trade-off \cite{CMIM,CMIM_WANG,mRR,Zhangis_1,DEAFS}. An effective way to achieve such trade-off is to evaluate feature dominance using Conditional Mutual Information (CMI) or Joint Mutual Information (JMI) \cite{JMI}, which are effective tools to simultaneously measure the class-relevance and the redundancy with one comprehensive value. Algorithms with Conditional Mutual Information Maximization (CMIM) criterion \cite{CMIM,CMIM_WANG} harness $\max_{F\in\s{F}} I(F ; C | \w{F})$ with $\w{F}$ $\st$ $\arg\min_{\w{F}\in \s{S}} I(F;C|\w{F})$ (where $I$ denotes CMI) to greedily select dominant features, where $\s{S}$ is the currently-selected subset and $\s{F}$ is the original feature set. A typical example of feature selection using JMI is 
the Double Input Symmetrical Relevance method (DISR) \cite{DISR}, which adopts the so called double input symmetrical relevance criterion to select features still in a greedy manner. DISR is in fact a modification of JMI and is shown to be one of the most effective MI-based feature selection criteria as indicated in \cite{FEAST}.

For the above MI-based feature selection criteria, it is noted that they all belongs to pairwise-approximation criteria, for they measure redundancy only relying on the magnitude of the dependency between any pairs of features rather than that among three or more features \cite{FEAST}. This heuristic significantly promotes executive efficiency and avoids the difficulty of sample inefficiency for joint distribution estimation \cite{mRR,Zhangis_1}. Whereas on the other hand, it inevitably leads to high-order information loss and thus may finally impair the quality of the selected features. The pros and cons of this heuristic are now still hot topics in MI-based feature selection research \cite{cluster_based,Zhangis_1}. Recently, Brown et al. \cite{FEAST} indicate that these pairwise-approximation criteria, such as MIM, MIFS, MIFS-U \cite{MIFS-U}, JMI, mRMR, and DISR, etc., can all be derived from a unified MI-based pairwise-approximation framework, with the criterion function
\begin{equation}\label{unified}
	J(F)=\alpha\cdot I(F;C)-\beta\cdot \sum_{F_s\in\s{S}}I(F;F_s)+\gamma\cdot\sum_{F_s\in\s{S}}I(F;F_s|C)
\end{equation}
where $\alpha$, $\beta$ and $\gamma$ are parameters of which any changes may correspond to an entirely new MI-based feature selection method. For example, MIM corresponds to $\alpha = 1$, $\beta = 0$ and $\gamma=0$, mRMR corresponds to $\alpha = 1$, $\beta = \frac{1}{|\s{S}|}$ and $\gamma = 0$, and DISR corresponds to $\alpha = \frac{1}{H(FF_sC)}$, $\beta=\frac{1}{|\s{S}|\cdot H(FF_sC)}$, and $\gamma=-\frac{1}{|\s{S}|\cdot H(FF_sC)}$, where $H(FF_sC)$ denotes the joint information entropy of $F,F_s,$ and $C$ (one may refer to \cite{FEAST} for detail). Due to more than one criterion as well as parameter to be considered and utilized (e.g. each item in the right side of Eq.\eqref{unified}), feature selection can thus be categorized as a multi-index evaluation process.

Data Envelopment Analysis (DEA) is an effective nonparametric method for efficiency evaluation and has been widely applied in many industries \cite{thirty_years_on}. It employs linear programming to evaluate and rank the Decision Making Units (DMUs) when the production process presents a structure of multiple inputs and outputs. Combining pattern recognition and data mining techniques with DEA has attracted increasing attention in recent decades, where DEA is often applied to evaluate the effectiveness of learning algorithms \cite{Ensembles_DEA,DEA_Classifier_1,DEA_Inter_Classifier,DEA_Classifier_2,cluster_based}. Zheng and Padmanabhan \cite{Ensembles_DEA} use DEA to construct an ensemble of classifiers in order to get better classification performance, which is a typical application of DEA to model combination problems. In addition, DEA itself is also applied to construct classifiers and clustering methods in prior work \cite{DEA_Classifier_1,DEA_Classifier_2,cluster_based}. Recently, Zhang et al. \cite{DEAFS} focus on the integration of DEA and feature selection, and indicate that DEA can be applied as a feature selection framework due to its nature of multi-index evaluation. It has the ability to give a reasonable overall score among more than one metric or criterion rather than only doing simple arithmetic operations among them, e.g. to get an overall score of a feature by only summing up the relevance score and the redundant one of this feature like what mRMR and DISR do. However, it has two fatal drawbacks: (1) it violates the rule-of-thumb in DEA since it conducts redundancy analysis by considering the candidate feature's conditional dependence to every other features in the feature space, and (2) such a redundancy analysis strategy considers both useful  features as well as useless ones which may obstacle the evaluation of conditional dependence and thus impair the quality of selected features.

In this paper, we still take feature selection as a multi-index evaluation process, and propose an effective feature selection method based on Class-Separability (CS) strategy and DEA, which can also effectively takle the problems of DEAFS. The unique contributions that distinguish our work from extant research are twofold:
\begin{itemize}
\item Class labels are taken as individual variables and CS strategy is then applied to explicitly handle relevance and redundancy on them, and
\item super-efficiency DEA is employed to evaluate and rank features via their conditional dependence scores on all class labels.
\end{itemize}
The remainder of the paper is organized as follows. Section \ref{metric} introduces related information-theoretic metrics, DEA, and the CS strategy. Section \ref{algo} proposes a novel method that integrates CS strategy and super-efficiency DEA into feature selection process. Then Section \ref{implementation} gives an implementation of the estimation of the measure applied in proposed method and analyzes the iteration and the computational complexity. In Section \ref{ExperimentalResults}, experimental results are given to evaluate the effectiveness of proposed method comparing with the representative feature selection methods on twelve well known UCI datasets, and some discussions are presented. Section \ref{conclusion} finally summarizes the concluding remarks and points out the future work.

\section{Foundations and basic concepts}\label{metric}
\subsection{Information theoretic metrics}
In information theory, Mutual Information (MI) is an essential metric of information and has been widely used for quantifying the mutual dependence of random variables. MI between two discrete random variables $X$ and $Y$, denoted as $I(X;Y)$, is formed as
\begin{equation*}
   I(X;Y)=
   \sum_{x\in X}\sum_{y\in Y}p(xy)\log\frac{p(xy)}{p(x)p(y)},
\end{equation*}
where $p(x)$ and $p(y)$ are the probabilities of the possible value assignments of $X$ and $Y$ respectively, and $p(xy)$ is the probability of the possible joint value assignments of $X$ and $Y$. Logarithms to the base 2 are used.\footnote{The notation log is used instead of $\log_2$ throughout this paper.}
MI measures dependence between $X$ and $Y$ in the following sense: $I(X; Y) = 0$ if and only if $X$ and $Y$ are independent random variables.

 Conditional Mutual Information (CMI) measures the conditional dependence between two random variables given the third. Given a discrete variable $Z$, CMI between $X$ and $Y$ is defined as
\begin{equation*}
  I(X;Y|Z)= \sum_{z\in Z}p(z)\sum_{x\in X}\sum_{y\in Y}p(xy|z)\log\frac{p(xy|z)}{p(x|z)p(y|z)}
\end{equation*}
$I(X;Y|Z)$ can be interpreted as the quantity of information that $X$ provides about $Y$ which $Z$ cannot. $I(X;Y|Z)=0$ implies no additional information $X$ can provide about $Y$ when the distribution of $Z$ is known, i.e. $X$ and $Y$ are conditional independent to each other given $Z$. Note that both MI and CMI are nonnegative ($\forall X,Y,Z, I(X;Y)\geq 0$ and $I(X;Y|Z)\geq 0$) and symmetric ($\forall X,Y,Z, I(X;Y)\equiv I(Y;X)$, $I(X;Y|Z)\equiv I(Y;X|Z)$).

\subsection{The super-efficiency Data Envelopment Analysis (DEA) model}\label{Framework}
In this section, we will briefly introduce the super-efficiency DEA model that will be used in our method. Before this, we first introduce some basic concepts of the basic DEA model originated by Charnes, Cooper and Rhodes (CCR model) \cite{CCR}. Consider $n$ Decision Making Units (DMUs) that use $m$ inputs to produce $s$ outputs. Let $x_{ij}$ $(i = 1,2,...,m)$ represents the $i$th input and $y_{rj}$ $(r = 1,2,...,s)$ the $r$th output of DMU$_j$ $(j = 1,2,...,n)$. Denote DMU$_p$ as the DMU that is under the evaluation, the relative efficiency of DMU$_p$ from CCR model can be achieved by solving the following linear programming
\begin{align}\label{CCR}
  \min_{\lambda_j,\theta_{p}^{c}} & \quad\hspace{2pt} {\theta}_{p}^c \\
  \st  & \quad \ds\sum_{j=1}^{n}{\lambda_j{x}_{ij}} \le {\theta}_{c}^p x_{ip},\quad i=1,2,...,m \nonumber \\
       & \quad \ds\sum_{j=1}^{n}{\lambda_j{y}_{rj}} \ge y_{rp},\quad r=1,2,...,s \nonumber\\
       & \quad \lambda_j\ge 0, \quad j=1,2,...,n \nonumber
\end{align}
where $\lambda_j$ is the coefficient corresponding to DMU$_j$. Denote the optimal value ${\theta_p^c}^*$ from model \eqref{CCR} as the relative efficiency of DMU$_p$, then DMU$_p$ is considered to be efficient if and only if ${\theta_{p}^{c}}^*=1$ (hereafter $\theta_{p}^*$ denotes optimal value for DMU$_p$); otherwise, it is referred to as non-efficient. However, CCR model can only separate DMUs into efficient ones (with the optimal value equal to 1) and inefficient ones (with the optimal value less than 1), for the objective value constraints to $0\le{\theta_p^c}^*\le 1$.

Super-efficiency DEA model \cite{thirty_years_on}, which is shown as
\begin{align}\label{Sup-CCR}
  \min_{\lambda_j,\theta_{p}^{s}} & \quad\hspace{2pt} {\theta}_{p}^s \\
  \st  & \quad \ds\sum_{j=1,j\ne p}^{n}{\lambda_j{x}_{ij}} \le {\theta}_{p}^s x_{ip},\quad i=1,2,...,m \nonumber \\
       & \quad \ds\sum_{j=1,j\ne p}^{n}{\lambda_j{y}_{rj}} \ge y_{rp},\quad r=1,2,...,s \nonumber\\
       & \quad \lambda_j\ge 0, \quad j=1,2,...,n,\quad j\ne p \nonumber
\end{align}
is one of the typical models that can fully rank DMUs. It is easy to find that ${\theta_{p}^{s}}^*\ge{\theta_{p}^{c}}^*$ since the optimal solutions from model \eqref{Sup-CCR} are always feasible to model \eqref{CCR}. The only difference between basic DEA and super-efficiency DEA is that the column in the data matrix which corresponds to the DMU-under-evaluation (i.e. DMU$_p$) is excluded from the matrix of the latter, i.e. the DMU-under-evaluation is excluded from the data envelopment and thus will not be used to construct the efficient frontier in super-efficiency DEA. For those DEA-inefficient DMUs, since they are under the efficient frontier, the frontier never changes no matter they are included in the data envelopment. Thus they will keep their efficiency score unchanged no matter using basic DEA or super-efficiency DEA, i.e. ${\theta_{p}^s}^* = {\theta_{p}^c}^*$ always holds for them. For those DEA-efficient DMUs, the efficient frontier may be changed because they locate on the original frontier. The different distances between the new frontier and the DEA-efficient DMUs result in the diversity of their efficiency scores. Thus, super-efficiency model is able to rank all the DMUs according to their efficiency scores.

\subsection{Class-Separability (CS) strategy}\label{CS-strategy}
As indicated in \cite{Zhangis_2}, most of the MI-based feature selection methods handle relevance and redundancy from a statistical perspective on class. That is, most of existing feature selection strategies assess feature(s) only on the `average' level of all class labels, but never directly capture the relationship between feature(s) and each class label. Since the classification accuracy will be finally reflected by both including the positive samples and excluding the negative ones for each class label, samples expressed by elaborately selected features will (hopefully) have strong links to their class labels. From this point of view, it is natural for feature selection methods to handle evaluation criteria (i.e. relevance and redundancy criteria) from the perspective of each class label. We call this Class-Separability (CS) strategy in feature selection, and we use the following example to illustrate it.
Given a feature $F$ and the class $C$, since $C$ always categorically ranges in the context of classification, $I(F;C)$ can thus be expressed as the following form

\begin{eqnarray}
				I(F;C) &=& \sum_{f\in F}\sum_{c\in C}p(fc)\log\frac{p(fc)}{p(f)p(c)} \nonumber\\
					   &=& \sum_{c\in C}p(c)\sum_{f\in F}p(f|c)\log\frac{p(f|c)}{p(f)} \nonumber\\
					   &=& \sum_{c\in C}p(c)Div(F;c) \label{Sep_c}
\end{eqnarray}
{where $Div(F;c)$ is the Kullback-–Leibler (KL) divergence of distribution $p(F)$ from conditional distribution $p(F|C=c)$ according to the information theory. Also, we may get
\begin{eqnarray}
				I(F;C|F_s) &=& \sum_{f\in F}\sum_{\s{f_s}\in \s{S}}\sum_{c\in C}p(f\s{f_s}c)\log\frac{p(fc|\s{f_s})}{p(f|\s{f_s})p(c|\s{f_s})} \nonumber\\
					   &=& \sum_{c\in C}p(c)\sum_{\s{f_s}\in \s{S}}p(\s{f_s})\sum_{f\in F}p(f|\s{f_s}c)\log\frac{p(f|\s{f_s}c)}{p(f|\s{f_s})} \nonumber\\
					   &=& \sum_{c\in C}p(c)Div(F;c|\s{S})  \label{Sep_cc}
\end{eqnarray}
where $\s{f_s}$ is the joint value assignment of the subset $\s{S}$. CS-strategy can thus be interpreted as  a multiple-objective programming of $Div(F;c)$ and $Div(F;c|\s{S})$ on all $c\in C$ rather than as a one-objective programming such as maximizing $I(F;C)$ and $I(F; C|\s{S})$ applied by traditional methods. In fact, all MI-based feature selection methods for classification can estimate (calculate) $I(F;C)$ and $I(F;C|\s{S})$ using Eqs.\eqref{Sep_c} -- \eqref{Sep_cc}. In other words, these methods explicitly separate the class into class labels and measure the relation from the perspective of each label during the estimation (calculation) process, but they neglect such separability and only take the weighted sum of them as an overall score and apply the score to handle relevance and redundancy. Decomposition of the estimation (calculation) process of MI and CMI can be shown as Fig.\ref{TraMethod}.

\vspace{1em}
\begin{figure*}[!ht]
\centering
\unitlength 1mm 
\linethickness{0.5pt}
\ifx\plotpoint\undefined\newsavebox{\plotpoint}\fi 
\begin{picture}(90,40)(0,0)
\put(22.75,24.25){\framebox(16,7.5)[cc]{label \#$1$}}
\put(22.75,15.5){\framebox(16,7)[cc]{label \#$2$}}
\put(22.75,1.5){\framebox(16,7.5)[cc]{label \#$k$}}
\put(39.25,12.5){\makebox(0,0)[cc]{......}}
\put(48.5,30.5){\makebox(0,0)[cc]{$Div(F,c_1|\s{S})$}}
\put(48.5,7.75){\makebox(0,0)[cc]{$Div(F,c_k|\s{S})$}}
\put(-5,15.25){\framebox(15,7.25)[cc]{Feature}}
\put(16,0){\framebox(49.95,40)[ct]{Class}}
\put(16,35){\line(1,0){49.95}}
\thicklines
\put(10.2,19){\vector(1,0){5.8}}
\put(22.75,27.5){\vector(3,4){.07}}\multiput(15.93,18.93)(.0330882,.0416667){17}{\line(0,1){.0416667}}
\multiput(17.055,20.346)(.0330882,.0416667){17}{\line(0,1){.0416667}}
\multiput(18.18,21.763)(.0330882,.0416667){17}{\line(0,1){.0416667}}
\multiput(19.305,23.18)(.0330882,.0416667){17}{\line(0,1){.0416667}}
\multiput(20.43,24.596)(.0330882,.0416667){17}{\line(0,1){.0416667}}
\multiput(21.555,26.013)(.0330882,.0416667){17}{\line(0,1){.0416667}}
\put(22.75,19){\vector(1,0){.07}}\put(15.93,18.93){\line(1,0){.9643}}
\put(17.858,18.93){\line(1,0){.9643}}
\put(19.787,18.93){\line(1,0){.9643}}
\put(21.715,18.93){\line(1,0){.9643}}
\put(22.584,4.492){\vector(1,-2){.07}}\multiput(15.997,18.927)(.031946,-.071105){12}{\line(0,-1){.071105}}
\multiput(16.763,17.221)(.031946,-.071105){12}{\line(0,-1){.071105}}
\multiput(17.53,15.514)(.031946,-.071105){12}{\line(0,-1){.071105}}
\multiput(18.297,13.808)(.031946,-.071105){12}{\line(0,-1){.071105}}
\multiput(19.064,12.101)(.031946,-.071105){12}{\line(0,-1){.071105}}
\multiput(19.83,10.395)(.031946,-.071105){12}{\line(0,-1){.071105}}
\multiput(20.597,8.688)(.031946,-.071105){12}{\line(0,-1){.071105}}
\multiput(21.364,6.982)(.031946,-.071105){12}{\line(0,-1){.071105}}
\multiput(22.13,5.275)(.031946,-.071105){12}{\line(0,-1){.071105}}
\put(66.1,18.577){\vector(1,0){.07}}\put(38.701,18.297){\line(1,0){.976}}
\put(40.653,18.312){\line(1,0){.976}}
\put(42.605,18.327){\line(1,0){.976}}
\put(44.557,18.342){\line(1,0){.976}}
\put(46.509,18.357){\line(1,0){.976}}
\put(48.461,18.372){\line(1,0){.976}}
\put(50.413,18.387){\line(1,0){.976}}
\put(52.366,18.402){\line(1,0){.976}}
\put(54.318,18.417){\line(1,0){.976}}
\put(56.27,18.432){\line(1,0){.976}}
\put(58.222,18.447){\line(1,0){.976}}
\put(60.174,18.462){\line(1,0){.976}}
\put(62.126,18.477){\line(1,0){.976}}
\put(64.078,18.492){\line(1,0){.976}}
\put(38.701,27.757){\line(1,0){.9737}}
\put(40.648,27.757){\line(1,0){.9737}}
\put(42.596,27.757){\line(1,0){.9737}}
\put(44.543,27.757){\line(1,0){.9737}}
\put(46.49,27.757){\line(1,0){.9737}}
\put(48.438,27.757){\line(1,0){.9737}}
\put(50.385,27.757){\line(1,0){.9737}}
\put(52.332,27.757){\line(1,0){.9737}}
\put(54.28,27.757){\line(1,0){.9737}}
\put(56.227,27.757){\line(1,0){.9737}}
\put(65.89,18.157){\vector(2,3){.07}}\multiput(57.201,5.052)(.0316882,.0479187){16}{\line(0,1){.0479187}}
\multiput(58.215,6.586)(.0316882,.0479187){16}{\line(0,1){.0479187}}
\multiput(59.229,8.119)(.0316882,.0479187){16}{\line(0,1){.0479187}}
\multiput(60.243,9.653)(.0316882,.0479187){16}{\line(0,1){.0479187}}
\multiput(61.257,11.186)(.0316882,.0479187){16}{\line(0,1){.0479187}}
\multiput(62.271,12.719)(.0316882,.0479187){16}{\line(0,1){.0479187}}
\multiput(63.285,14.253)(.0316882,.0479187){16}{\line(0,1){.0479187}}
\multiput(64.299,15.786)(.0316882,.0479187){16}{\line(0,1){.0479187}}
\multiput(65.313,17.32)(.0316882,.0479187){16}{\line(0,1){.0479187}}
\put(38.701,4.842){\line(1,0){.9626}}
\put(40.626,4.82){\line(1,0){.9626}}
\put(42.551,4.798){\line(1,0){.9626}}
\put(44.477,4.776){\line(1,0){.9626}}
\put(46.402,4.754){\line(1,0){.9626}}
\put(48.327,4.732){\line(1,0){.9626}}
\put(50.252,4.709){\line(1,0){.9626}}
\put(52.177,4.687){\line(1,0){.9626}}
\put(54.103,4.665){\line(1,0){.9626}}
\put(56.028,4.643){\line(1,0){.9626}}
\put(65.89,18.93){\vector(1,-1){.07}}\multiput(57,27.867)(.0331933,-.0347739){19}{\line(0,-1){.0347739}}
\multiput(58.252,26.645)(.0331933,-.0347739){19}{\line(0,-1){.0347739}}
\multiput(59.513,25.324)(.0331933,-.0347739){19}{\line(0,-1){.0347739}}
\multiput(60.775,24.003)(.0331933,-.0347739){19}{\line(0,-1){.0347739}}
\multiput(62.036,22.681)(.0331933,-.0347739){19}{\line(0,-1){.0347739}}
\multiput(63.297,21.36)(.0331933,-.0347739){19}{\line(0,-1){.0347739}}
\multiput(64.559,20.038)(.0331933,-.0347739){19}{\line(0,-1){.0347739}}
\put(48.5,20.679){\makebox(0,0)[cc]{$Div(F,c_2|\s{S})$}}
\put(65.95,18.367){\vector(1,0){9.04}}
\put(79.05,18.577){\circle{7.821}}
\put(78.88,18.577){\makebox(0,0)[cc]{$\sum$}}
\put(83.15,18.577){\vector(1,0){8.07}}
\put(100,18.577){\makebox(0,0)[cc]{$I(F,C|\s{S})$}}
\end{picture}
\caption{Extant MI-based feature relevance (redundancy) evaluation process, where $\sum$ denotes weighted summation operation. Note that $I(F;C|\s{S}) = I(F;C)$ if $\s{S}=\emptyset$.}
\label{TraMethod}
\end{figure*}
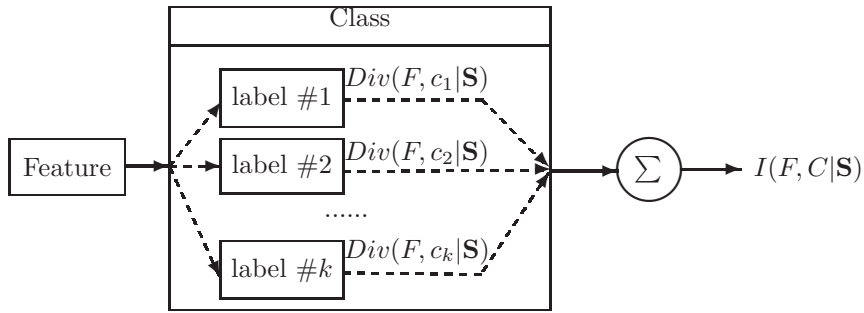
\vspace{1em}

Fig. \ref{TraMethod} indicates that, it is possible to explicitly use $Div(F;c_i|\s{S})$ rather than the average (conditional) mutual information $I(F;C|\s{S})$ to handle relevance and redundancy from the perspective of each class label, without changing the estimation (calculation) process of MI and CMI. We will use such CS strategy with MI and CMI to construct our method. 

\section{Evaluating and selecting features with CS Strategy and super-efficiency DEA}\label{algo}

In what follows, we introduce a novel feature evaluation and selection method which integrates CS strategy and super-efficiency DEA into its searching process. To identify salient features that have strong discriminative power and the power would not be impaired given other features, conditional dependence between the feature-under-evaluation and the class given the currently-selected subset is often harnessed to select features with a greedy search strategy \cite{IAMBs,IAMB2010,Zhangis_1}. That is, features are incrementally selected in the currently-selected subset via the following process
\begin{equation}\label{greedy}\s{S}_{k+1}=\s{S}_k\cup\left\{F_{k+1}\left|F_{k+1}=\arg\max_{F\in\s{F}-\s{S}_k}\text{CD}(F;C|\s{S}_k)\right\}\right.\end{equation}
where $\s{S}_k$ is the currently-selected subset obtained at the end of the $k$th step, $F_{k+1},\s{S}_{k+1}$ are the newly-selected feature and the currently-selected subset obtained at the end of the $k+1$th step, respectively. CD denotes the conditional dependence metric and it is often implemented with CMI \cite{IAMB2010,Zhangis_1}. The metric measures conditional dependence between two variables with respect to the class concept: A very small value of $\text{CD} (F; C | \s{S})$ implies that $F$ carries little additional information about $C$ given the subset $\s{S}$, namely (a) $F$ is redundant to $\s{S}$, or (b) $F$ is irrelevant to $C$. The advantage of spontaneously paying attention to both relevance and redundancy makes conditional dependence metric frequently used in the related work \cite{CMIM,CMIM_WANG,Zhangis_1}.

However, as we mentioned in subsection \ref{CS-strategy}, the conditional dependence metric such as CMI takes both the feature and the class as individual random variables and measures the magnitude of the relation between them from a statistical perspective. Whereas in a classification task, the feature and the class are obviously not in the same position (e.g. we use value assignments of features to predict class labels), and the classification accuracy is finally reflected by both including the positive samples and excluding the negative ones for each class label. In addition, features with high class-relevance (low redundancy), i.e. with large conditional dependence scores, may not be always strongly relevant to (significantly non-redundant on) all class labels. For example, according to Eqs.\eqref{Sep_c} and \eqref{Sep_cc}, the feature with maximum mutual information $I(F;C)$ or conditional mutual information $I(F;C|\s{S})$ among all candidates can not guarantee maximum $Div(F;c)$ or $Div(F;c|\s{S})$ for all $c\in C$, respectively. This implies that features usually do not have the same discriminative power for different class labels. Therefore, it may turn out to be that ignorance of the conditional dependence distribution on the class finally impairs the discriminative power of the selected feature subset. To this end, CS strategy is thus applied in our method to explicitly handle relevance and redundancy on each class label. To evaluate features with respect to the conditional dependence distribution (i.e. the histogram with more than one value since there are more than one class label), feature selection is thus taken as a multi-index evaluation process and DEA technique is harnessed to evaluate features with respect to the conditional dependence distribution, rather than only summing up $p(c)\cdot Div(F;c)$ and $p(c)\cdot Div(F;c|\s{S})$ like what traditional MI-based methods do (Fig. \ref{TraMethod}). We show this process in Fig. \ref{ProposedFramework}.

\begin{figure*}[!ht]
\vspace{1em}
\centering
\unitlength 1mm 
\linethickness{0.5pt}
\ifx\plotpoint\undefined\newsavebox{\plotpoint}\fi 
\begin{picture}(90,40)(0,0)
\put(22.75,24.25){\framebox(16,7.5)[cc]{label \#$1$}}
\put(22.75,15.5){\framebox(16,7)[cc]{label \#$2$}}
\put(22.75,1.5){\framebox(16,7.5)[cc]{label \#$k$}}
\put(39.25,12.5){\makebox(0,0)[cc]{......}}
\put(48,30.5){\makebox(0,0)[cc]{$R(F,c_1|\s{S})$}}
\put(47.5,7.75){\makebox(0,0)[cc]{$R(F,c_k|\s{S})$}}
\put(-5,15.25){\framebox(15,7.25)[cc]{Feature}}
\put(16,0){\framebox(49.95,40)[ct]{Class}}
\put(16,35){\line(1,0){49.95}}
\thicklines
\put(10.2,19){\vector(1,0){5.8}}
\put(22.75,27.5){\vector(3,4){.07}}\multiput(15.93,18.93)(.0330882,.0416667){17}{\line(0,1){.0416667}}
\multiput(17.055,20.346)(.0330882,.0416667){17}{\line(0,1){.0416667}}
\multiput(18.18,21.763)(.0330882,.0416667){17}{\line(0,1){.0416667}}
\multiput(19.305,23.18)(.0330882,.0416667){17}{\line(0,1){.0416667}}
\multiput(20.43,24.596)(.0330882,.0416667){17}{\line(0,1){.0416667}}
\multiput(21.555,26.013)(.0330882,.0416667){17}{\line(0,1){.0416667}}
\put(22.75,19){\vector(1,0){.07}}\put(15.93,18.93){\line(1,0){.9643}}
\put(17.858,18.93){\line(1,0){.9643}}
\put(19.787,18.93){\line(1,0){.9643}}
\put(21.715,18.93){\line(1,0){.9643}}
\put(22.584,4.492){\vector(1,-2){.07}}\multiput(15.997,18.927)(.031946,-.071105){12}{\line(0,-1){.071105}}
\multiput(16.763,17.221)(.031946,-.071105){12}{\line(0,-1){.071105}}
\multiput(17.53,15.514)(.031946,-.071105){12}{\line(0,-1){.071105}}
\multiput(18.297,13.808)(.031946,-.071105){12}{\line(0,-1){.071105}}
\multiput(19.064,12.101)(.031946,-.071105){12}{\line(0,-1){.071105}}
\multiput(19.83,10.395)(.031946,-.071105){12}{\line(0,-1){.071105}}
\multiput(20.597,8.688)(.031946,-.071105){12}{\line(0,-1){.071105}}
\multiput(21.364,6.982)(.031946,-.071105){12}{\line(0,-1){.071105}}
\multiput(22.13,5.275)(.031946,-.071105){12}{\line(0,-1){.071105}}
\put(66.1,18.577){\vector(1,0){.07}}\put(38.701,18.387){\line(1,0){.976}}
\put(40.653,18.387){\line(1,0){.976}}
\put(42.605,18.387){\line(1,0){.976}}
\put(44.557,18.387){\line(1,0){.976}}
\put(46.509,18.387){\line(1,0){.976}}
\put(48.461,18.387){\line(1,0){.976}}
\put(50.413,18.387){\line(1,0){.976}}
\put(52.366,18.387){\line(1,0){.976}}
\put(54.318,18.387){\line(1,0){.976}}
\put(56.27,18.387){\line(1,0){.976}}
\put(58.222,18.387){\line(1,0){.976}}
\put(60.174,18.387){\line(1,0){.976}}
\put(62.126,18.387){\line(1,0){.976}}
\put(64.078,18.387){\line(1,0){.976}}

\put(38.701,27.757){\line(1,0){.9737}}
\put(40.648,27.757){\line(1,0){.9737}}
\put(42.596,27.757){\line(1,0){.9737}}
\put(44.543,27.757){\line(1,0){.9737}}
\put(46.49,27.757){\line(1,0){.9737}}
\put(48.438,27.757){\line(1,0){.9737}}
\put(50.385,27.757){\line(1,0){.9737}}
\put(52.332,27.757){\line(1,0){.9737}}
\put(54.28,27.757){\line(1,0){.9737}}
\put(56.227,27.757){\line(1,0){.9737}}
\put(65.89,18.157){\vector(2,3){.07}}\multiput(57.201,5.052)(.0316882,.0479187){16}{\line(0,1){.0479187}}
\multiput(58.215,6.586)(.0316882,.0479187){16}{\line(0,1){.0479187}}
\multiput(59.229,8.119)(.0316882,.0479187){16}{\line(0,1){.0479187}}
\multiput(60.243,9.653)(.0316882,.0479187){16}{\line(0,1){.0479187}}
\multiput(61.257,11.186)(.0316882,.0479187){16}{\line(0,1){.0479187}}
\multiput(62.271,12.719)(.0316882,.0479187){16}{\line(0,1){.0479187}}
\multiput(63.285,14.253)(.0316882,.0479187){16}{\line(0,1){.0479187}}
\multiput(64.299,15.786)(.0316882,.0479187){16}{\line(0,1){.0479187}}
\multiput(65.313,17.32)(.0316882,.0479187){16}{\line(0,1){.0479187}}
\put(38.701,4.842){\line(1,0){.9626}}
\put(40.626,4.82){\line(1,0){.9626}}
\put(42.551,4.798){\line(1,0){.9626}}
\put(44.477,4.776){\line(1,0){.9626}}
\put(46.402,4.754){\line(1,0){.9626}}
\put(48.327,4.732){\line(1,0){.9626}}
\put(50.252,4.709){\line(1,0){.9626}}
\put(52.177,4.687){\line(1,0){.9626}}
\put(54.103,4.665){\line(1,0){.9626}}
\put(56.028,4.643){\line(1,0){.9626}}
\put(65.89,18.787){\vector(1,-1){.07}}\multiput(57,27.867)(.0331933,-.0347739){19}{\line(0,-1){.0347739}}
\multiput(58.252,26.645)(.0331933,-.0347739){19}{\line(0,-1){.0347739}}
\multiput(59.513,25.324)(.0331933,-.0347739){19}{\line(0,-1){.0347739}}
\multiput(60.775,24.003)(.0331933,-.0347739){19}{\line(0,-1){.0347739}}
\multiput(62.036,22.681)(.0331933,-.0347739){19}{\line(0,-1){.0347739}}
\multiput(63.297,21.36)(.0331933,-.0347739){19}{\line(0,-1){.0347739}}
\multiput(64.559,20.038)(.0331933,-.0347739){19}{\line(0,-1){.0347739}}
\put(47.811,20.679){\makebox(0,0)[cc]{$R(F,c_2|\s{S})$}}
\put(65.95,18.367){\vector(1,0){9.04}}
\put(79.05,18.577){\circle{7.821}}
\put(78.88,18.577){\makebox(0,0)[cc]{DEA}}
\put(83.15,18.577){\vector(1,0){8.07}}
\put(96,18.577){\makebox(0,0)[cc]{$\theta_{F}^*$}}
\end{picture}
\caption{Feature relevance (redundancy) evaluation process with class-separability strategy and DEA.}
\label{ProposedFramework}
\vspace{1em}
\end{figure*}
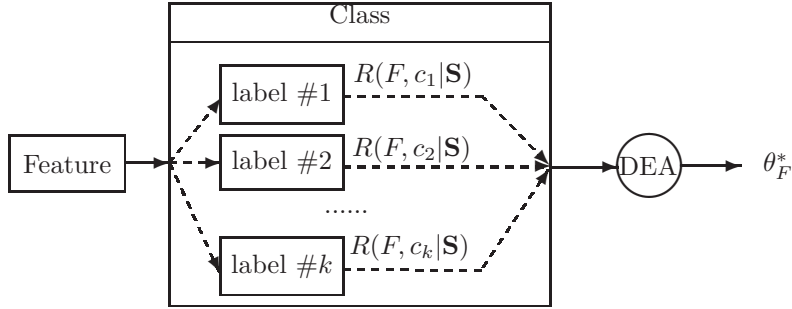

To better capture the relationship between the feature and each class label, we take each class label as an individual variable. In other words, we measure MI (CMI) between $F$ and $c_i$ by both considering $Div(F;c_i)$ $(Div(F;c_i|\s{S}))$ and $Div(F;\bar{c}_i)$ $(Div(F;\bar{c}_i|\s{S}))$, where $\bar{c}_i$ is an artificial class label on behalf of all other class labels in $C$ except $c_i$. We thus define the conditional independence score of feature $F$ given the feature subset $\s{S}$ as
\begin{equation}\label{R}
  R(F;c_i|\s{S}) = p(c_i)\cdot Div(F;c_i|\s{S})+p(\bar{c}_i)\cdot Div(F;\bar{c}_i|\s{S}).
\end{equation}
To illustrate Eq.\eqref{R}, we set $\s{C}=\{C_1,...,C_k\}$ where $C_i=\{c_i,\bar{c}_i\}$ $(i=1,...,k)$. Thus, $C_i$ is the new class that can take place of $C$ when only caring the class label $c_i$ on the whole samples. According to Eq.\eqref{Sep_cc}, CMI between $F$ and $C_i$ is expressed as
\begin{eqnarray}
	I(F;C_i|\s{S}) 
	&=& \sum_{c\in C_i}p(c)Div(F;c|\s{S}) \nonumber \\
	               &=& p(c_i)\cdot Div(F;c_i|\s{S})+p(\bar{c}_i)\cdot Div(F;\bar{c}_i|\s{S})\nonumber
\end{eqnarray}
Hence we get the form of Eq.\eqref{R}. The underlying intention of Eq.\eqref{R} is (a) to measure MI (CMI) between two variables rather than that between a variable and the value of a class label which cannot capture the information about the absence of that class label, (b) to partially mitigate the negative effects of sample inefficiency problem since the magnitude of the relation between a feature and a class label is measured on the whole samples rather than only on the samples with that class label, and (c) to keep the score always nonnegative whereas $Div(F;c)$ $(Div(F;c|\s{S}))$ cannot, which satisfies the nonnegative constraint in DEA (e.g. the last constraint shown in models \eqref{CCR} and \eqref{Sup-CCR}).

We now obtain a feature evaluation system with $|C|$ outputs related to conditional dependence for all candidate features. Since the magnitudes of the outputs (i.e. the values of $R(F;c_i|\s{S})$ for all $c_i$) have strong influence on the quality of the selected features, we set the input to a constant for all candidate features to control the problem scale. Thus, the feature evaluation system involves $|C|$ outputs and a constant input, and each candidate feature is taken as a DMU waiting for evaluation (Fig. \ref{DEAframework}). In order to fully rank all candidate features, a super-efficiency DEA model is applied to execute the evaluation task of the system, which is shown as model \eqref{core}. We apply the following greedy search strategy to select features 
\begin{equation}\label{Ourgreedy}\s{S}_{k+1}=\s{S}_k\cup\left\{F_{k+1}\left|F_{k+1}=\arg\max_{F\in\s{F}-\s{S}_k}\theta_{F}^{s*}\right\}\right.\end{equation}
 where $\theta_{F}^{s*}$ is achieved by solving Model \eqref{core}.
\begin{figure*}[!ht]
\centering
\unitlength 1mm 
\linethickness{0.5pt}
\ifx\plotpoint\undefined\newsavebox{\plotpoint}\fi 
\begin{picture}(80,18)(0,0)
\put(15.41,0){\framebox(36.75,18)[cc]{$F_p$ (DMU$_p$)}}
\put(52.326,10.875){\vector(1,0){12.021}}
\put(52.326,2){\vector(1,0){12.021}}
\put(73.362,10.625){\makebox(0,0)[cc]{$R(F_p,c_2|\s{S})$}}
\put(73.362,14.868){\makebox(0,0)[cc]{$R(F_p,c_1|\s{S})$}}
\put(73.362,1.556){\makebox(0,0)[cc]{$R(F_p,c_k|\s{S})$}}
\put(58.16,6.029){\makebox(0,0)[cc]{$......$}}
\put(52.326,15.65){\vector(1,0){12.021}}
\put(3.536,9.10){\vector(4,0){11.844}}
\put(0,9.211){\makebox(0,0)[cc]{$M$}}
\end{picture}
\caption{Proposed feature evaluation system with a constant input and CMI values of all class labels as outputs.}
\label{DEAframework}
\end{figure*}
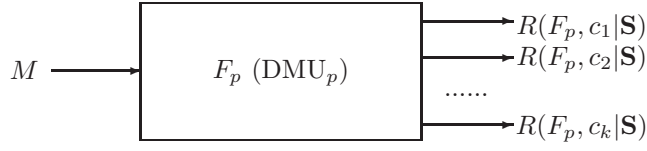

\begin{align}\label{core}
  \min_{\lambda_j,\theta^{s}_{F_p}} & \quad\hspace{2pt} {\theta}^{s}_{F_p} \\
  \st  & \quad \sum_{j=1,j\ne p}^{|\s{\Omega}|}\lambda_j\cdot R(F_j,c_r|\s{S}) \ge R(F_p,c_r|\s{S}), \quad r =  1,2,...,|C| \nonumber \\
       & \quad \sum_{j=1,j\ne p}^{|\s{\Omega}|}\lambda_j \le {\theta}^{s}_{F_p}\nonumber\\
       & \quad \lambda_j\ge 0, \quad j=1,2,...,|\s{\Omega}|,\quad j\ne p. \nonumber
\end{align}
Note that the second constraint in model \eqref{core}, i.e. $\sum_{j=1,j\ne p}^{|\s{\Omega}|}\lambda_j \le {\theta}^{s}_{F_p}$, is the simplification of the following inequality
\begin{equation}
\sum_{j=1,j\ne p}^{|\s{\Omega}|}\lambda_j\cdot M \le {\theta}^{s}_{F_p}\cdot M
\end{equation}
which is the standard form of the constraint for the input of DMUs. Model \eqref{core} does not care the inputs but only focuses on the magnitude as well as the diversity of the outputs and hence can avoid the weakness that DEA excessively focuses on the `gaps' between the inputs and the outputs rather than the magnitudes of the outputs. This weakness makes DEA not so effective sometimes in feature selection since the magnitude of the outputs (e.g. class-relevance of the feature) often influences the quality of selected features \cite{DEAFS}.  

\begin{algorithm}[!th]
   \caption{DEA-CS: Feature selection algorithm with super-efficiency{\bf DEA} and {\bf C}lass-{\bf S}eparability strategy}
   \label{alg}
   \KwIn{ $\s{D}$ /*dataset*/, $\s{F}$ /*feature set*/, $C$ /*class*/, $\delta$ /*expected \# features to be selected*/}
   \KwOut{$\s{S}$ /*selected feature subset*/}
   Initialize $\s{S}\leftarrow\varnothing$, $k\leftarrow 1$  \\
     \Repeat{$k > \delta$}{
       $\s{\Gamma}\leftarrow \varnothing$,
       $\Sigma\leftarrow 0$   \\
	   \ForEach{$F\in\s{F}-\s{S}$}{
	       $\sigma\leftarrow 0$ \\
	   	   \For{$i=1$ to $|C|$}{
	           $\Gamma\leftarrow \Gamma\cup \{R(F;c_i|\s{S})\}$,
	           $\sigma\leftarrow\sigma+R(F;c_i|\s{S})$
	       }
	       \If{$\sigma = 0$}{\bf{continue}}
	       $\s{\Gamma}\leftarrow\s{\Gamma}\cup\{\Gamma\}$,$\Gamma\leftarrow\varnothing$  \\
	       $\Sigma\leftarrow\Sigma+\sigma$  \\
	   }
	   \If{$\Sigma = 0$}{\bf{break}}
	   $\theta_{max}\leftarrow$SupDEAsolver$(\s{\Gamma})$ \\
	   $\s{S}\leftarrow\s{S}\cup\left\{ F_p\right\}$ $\st$ $F_p={\displaystyle\arg\theta_{max}}$ \\
	   $k\leftarrow k + 1$
   }
   \Return{$\s{S}$}
\end{algorithm}

The pseudo code of DEA-CS shown in Algorithm \ref{alg} contains a `repeat' loop and two `for' loops. The `repeat' loop (steps 2-21) circulates $\delta$ times, but it will stop once $\Sigma=0$ (step 15). In fact, sample inefficiency often makes the estimation of conditional mutual information be 0 (which results in $\Sigma=0$) even with a medium-scale conditioning set $\s{S}$, so the `repeat' loop circulates with the worst-case time complexity of $\mathrm{O}(\delta)$. As for the two `for' loops, the outer one (steps 4-14) totally circulates with the worst-case time complexity of $\mathrm{O}(|\s{F}|)$, while the inner one (steps 6-11) circulates with the time complexity of $\mathrm{\Theta}(|C|)$. Consequently, the worst-case iteration complexity of DEA-CS is $\mathrm{O}(\delta\cdot|\s{F}|\cdot|C|)$. Line 18 in Algorithm \ref{alg} solves super-efficiency DEA (model \eqref{core}) and gets a maximum efficiency score among all candidate features. The process of SupDEAsolver($\cdot$) in line 18 is given as Algorithm \ref{supDEAsol}.
\begin{algorithm}[!th]
   \caption{SupDEAsolver: To solve the super-efficiency DEA problem with Eq.\eqref{core}.}
   \label{supDEAsol}
   \KwIn{ $\s{\Gamma}$ /*set of the outputs for solving Eq.\eqref{core}*/}
   \KwOut{$\theta_{max}$ /*maximum efficiency score*/}
   Initialize $\theta_{max}\leftarrow0$  \\
	   \ForEach{$p=1$ to $ |\s{\Gamma}|$}{
	      Take the elements in $\Gamma_p$ as the outputs of DMU$_p$ and the elements in $\Gamma_j\in\s{\Gamma}$ as the outputs of DMU$_j (j=1,...,|\s{\Gamma}|, j\ne p)$ \\
	      Solve model \eqref{core} to get the efficiency score ${\theta_p^s}^*$ \\
	      \If{$\theta_{max}\le{\theta_p^s}^*$}{
	          $\theta_{max} \leftarrow {\theta_p^s}^*$
	      }
	   }
   \Return{$\theta_{\text{max}}$}
\end{algorithm}

\section{Algorithm implementation and complexity analysis}\label{implementation}
\subsection{Estimation of $R(F;c|\s{S})$}
In order to achieve the computational complexity of DEA-CS, we first analyze the estimation process of $R(F,c|\s{S})$. Before this, we first introduce the definition of Local Mutual Information (LMI) as follows. Note that the estimation process in proposed algorithm is based on that in CCM \cite{Zhangis_1} which only considers the value assignments existing in the samples.

\begin{define}[{\bf{Local Mutual Information }}\cite{Zhangis_1}]
\label{LMI}
    Suppose that $\s{D}$ is the sample set which contains $N$ samples, $\w{\s{D}}$ is the sample subset $\st$ $\w{\s{D}}\subset\s{D}$, then the estimated local mutual information between two discrete features $X$ and $Y$ on $\w{\s{D}}$ is defined as
    \begin{equation}\label{LMIeqn}
        \hat{I}_{\w{\s{D}}}(X;Y) = \sum_{x\in X}\sum_{y\in Y}\hat{p}_{\w{\s{D}}}(xy)\log\frac{\hat{p}_{\w{\s{D}}}(xy)}{\hat{p}_{\w{\s{D}}}(x)\hat{p}_{\w{\s{D}}}(y)},
    \end{equation}
    where $\hat{p}_{\w{\s{D}}}(\cdot)$ denote probability or joint probability estimated on $\s{\w{D}}$.
\end{define}
With the definition of LMI, we have the following proposition.

\begin{prop}[\cite{Zhangis_1}]
\label{LMI&CMI}
   Suppose there are in total $m$ different joint value assignments of $\s{S}$ in $\s{D}$, then the estimated CMI between $X$ and $Y$ given $\s{S}$ can be expressed as
    \begin{eqnarray}
    \label{LMI&CMI_eqn}
        \hat{I}(X;Y|\s{S})=\sum_{i=1}^{m}\frac{|\s{D}_i|}{|\s{D}|} \hat{I}_{\s{D}_i}(X;Y),
    \end{eqnarray}
    where $\s{D}_j$ is the sample subset in which samples contain the $i$th joint value assignment of $\s{S}$, and $\displaystyle \bigcup_{j=1}^{m}\s{D}_j\subset\s{D}$.
\end{prop}

With Proposition \ref{LMI&CMI}, it is straightforward that $R(F;c_i|\s{S})$ can be estimated similarly with $I(F;C|\s{S})$. In terms of \cite{Zhangis_1}, the estimation of $R(F;c_i|\s{S})$ involves two steps: sample sorting and LMI estimation. That is, samples should be first ranked with respect to the joint value assignments of $\s{S}$ in order to make $\s{D}_j (j=1,...,m)$ satisfy 
\begin{equation}\label{arranged_1}  \bigcup_{j=1}^m\s{D}_j = \s{D} \end{equation} 
and
\begin{equation}\label{arranged_2}  \bigcap_{j=1}^{m}\s{D}_j = \emptyset.\end{equation}
To realize this, sorting methods such as merge-sort method (with the computational complexity of $\mathrm{O}(|\s{S}|\cdot|\s{D}|\cdot\log|\s{D}|)$) and radix-sort method (with the computational complexity of $\mathrm{O}(|\s{S}|\cdot(|\s{D}|+\bar{r}))$ where $\bar{r}= \max_{F\in\s{S}}|F|$) \footnote{$|F|$ denotes \# value assignments of F in $\s{D}$} can be applied. Then, LMI estimation with linear computational complexity can be conducted on each $\s{D}_j\subset \s{D}$ by traversing $\s{D}$ only once. Since the joint distribution $p(fc_i)$ for all pairs of $f$ and $c_i$ have already achieved during the traversing process, all $R(F;c_i|\s{S})$ ($c_i\in C$) can be estimated within the complexity of $\mathrm{O}(|\s{S}|\cdot|\s{D}|\cdot\log|\s{D}|)$ (if merge-sort method is applied) or $\mathrm{O}(|\s{S}|\cdot(|\s{D}|+\bar{r}))$ (if radix-sort method is applied)

More specifically, with the characteristic of DEA-CS that the scale of the selected feature subset $\s{S}$ grows incrementally, we have the following corollary.
\begin{corollary}\label{coro}
	Given the current feature subset $\s{S}_{t}$ and the newly-selected feature $F_{added}$, the wost-case computational complexity of the estimation of all $R(F;c_i|\s{S}_{t+1})$ ($i=1,...,|C|$) in DEA-CS is $\mathrm{O}(|\s{D}|\cdot\log|\s{D}|)$ or $\mathrm{O}(|\s{D}|+\bar{r})$ where $\bar{r} = |F_{added}|$.
\end{corollary}
Proof of Corollary \ref{coro} is straightforward, since there is only one feature $F_{added}$ which will be selected and added in $\s{S}_t$ at the $t$th iteration of the first `for' loop of DEA-CS shown in Algorithm \ref{alg}. Thus, we only need to re-sort samples with respect to the value assignments of $F_{added}$ to get $\s{S}_{t+1}$ with well-ordered samples satisfying Eq.\eqref{arranged_1} and Eq.\eqref{arranged_2}, with the computational complexity of $\mathrm{O}(|\s{D}|\cdot\log|\s{D}|)$, or $\mathrm{O}(|\s{D}|+\bar{r})$ where $\bar{r} = |F_{added}|$. Under this circumstance, the estimation of all $R(F;c_i|\s{S}_{t+1})$ ($i=1,...,|C|$) (i.e. LMI estimation) can be conducted with the computational complexity of $\mathrm{O}(|\s{D}|)$. Therefore, the wost-case computational complexity of the estimation of all $R(F;c_i|\s{S}_{t+1})$ ($i=1,...,|C|$) in DEA-CS is $\mathrm{O}(|\s{D}|\cdot\log|\s{D}|)$ or $\mathrm{O}(|\s{D}|+\bar{r})$ where $\bar{r} = \max_{F\in\s{F}}|F|$.

\subsection{Computational complexity of DEA-CS}
The computational complexity of DEA-CS refers to the estimation of $R(F,c|\s{S})$ $(c\in C)$ and linear programming (LP) solving, while that of the former has been given in Corollary \ref{coro}. For LP solving, interior methods can be applied in DEA-CS to guarantee polynomial time complexity, e.g. Karmarkar's famous algorithm \cite{karmarkar} with the computational complexity of $\mathrm{O}(|\s{D}|^{3.5} L)$, where $L$ is the bi-length of data. With the results of the iteration complexity of DEA-CS, the computational complexity of the estimation of $R(F,c|\s{S}) for all c\in C$, and the computational complexity of a LP solver, we can thus get the computational complexity of DEA-CS as $\mathrm{O}(\delta\cdot|\s{F}|\cdot|\s{D}|\cdot\log|\s{D}|)$ or $\mathrm{O}(\delta\cdot|\s{F}|\cdot(|\s{D}|+\bar{r}))$. Note that in practice, interior methods may not be necessarily used for DEA solving in DEA-CS since simplex method may be more efficient if the problem scale is not large. 

\section{Empirical results}\label{ExperimentalResults}
\label{ExpRes}
In this section we empirically evaluate proposed method against four representative feature selection methods. Ten well-known UCI datasets \footnote{\url{http://archive.ics.uci.edu/ml/}} are chosen in our experiments for performance validation. In what follows, we first briefly describe
the benchmark datasets and the experiment settings. The experimental results are then presented and discussed.

\subsection{Datasets and experimental settings}
Table \ref{datasets} summarizes general information about the selected UCI datasets. These are chosen from different domains and to have a wide range of multi-class problems. The features within
each dataset have a variety of characteristics --- some discrete and some continuous. Continuous features are discretized using Minimum Descriptive Length (MDL) method \cite{MDL}, which is a supervised discretization method, prior to feature selection and classification. Features already with a discrete range are left untouched.
\begin{table}[!ht]
\centering
    \begin{threeparttable}
    \caption{Datasets used in the experiments.}\label{datasets}
        \begin{tabular}{rl r@{}lr@{}lcl}
            \toprule
                 Id\phantom{.000}  & Name  & \multicolumn{2}{l}{\phantom{}\# samples\phantom{00}} & \multicolumn{2}{l}{\phantom{}\# features\phantom{0}} & Type & \# classes \\
            \midrule
           1\phantom{0000}&     flags       &  \phantom{00} 194     &    &  \phantom{00} 29     &  & discrete+continuous  &  \phantom{0000}8\\
           2\phantom{0000}&     kr-vs-kp    &  \phantom{00} 3196    &    &  \phantom{00} 36     &  & discrete  &  \phantom{0000}2\\
           3\phantom{0000}& mfeat-zernike   &  \phantom{00} 2000    &    &  \phantom{00} 48     &  & continuous  &  \phantom{000}10\\
           4\phantom{0000}&synthetic-control&  \phantom{00} 600     &    &  \phantom{00} 60     &  & continuous  &  \phantom{0000}6\\
           5\phantom{0000}& splice          &  \phantom{00} 3190    &    &  \phantom{00} 60     &  & discrete  &  \phantom{0000}3\\
           6\phantom{0000}&     musk2       &  \phantom{00} 6598    &    &  \phantom{00} 166    &  & continuous  &  \phantom{0000}2\\
           7\phantom{0000}&     DNA         &  \phantom{00} 3186    &    &  \phantom{00} 180    &  & discrete  &  \phantom{0000}3\\
           8\phantom{0000}& arrhythmia      &  \phantom{00} 452     &    &  \phantom{00} 279    &  & discrete+continuous  &  \phantom{000}13\\
          9\phantom{0000}&     isolet5     &  \phantom{00} 1559    &    &  \phantom{00} 617    &  & continuous  &  \phantom{000}26\\
          10\phantom{0000}&     gisette     &  \phantom{00} 6000    &    &  \phantom{00}5000    &  & discrete+continuous  &  \phantom{0000}2\\
            \bottomrule
        \end{tabular}
    \end{threeparttable}
\end{table}

Since the classification performance is the ultimate evaluation for feature selection validation \cite{FEAST}, we report in our experiments the classification results on selected UCI data sets. Four typical classifiers, namely Support Vectoer Machine (SVM) \cite{SVM}, C4.5 decision tree \cite{5}, $k$-Nearest Neighbor ($k$NN) \cite{kNN}, and Na\"{i}ve Bayesian Classifier (NBC) \cite{Weka}, are selected to perform classification task. They are the most influential algorithms that have been widely used in the pattern recognition and data mining communities. The experimental workbench for classification is Weka (Waikato environment for knowledge analysis) \cite{Weka}, which is a collection of typical data mining algorithms. The parameters of classifiers for each experiment are set to default values preset in Weka. We use in our experiments classification accuracy as the performance index.

we empirically evaluate the performance of proposed method by comparing it with three MI-based feature selection methods namely mRMR \cite{9}, DISR \cite{DISR}, and MIM \cite{MIM}, a famous distance-based method ReliefF \cite{ReliefF}, and another DEA-based feature ranking algorithm DEAFS \cite{DEAFS}. All of them are the most representative and well-performed feature selection methods. A brief description of them is as follows:
\begin{itemize}
\item mRMR algorithm \cite{9}: The minimal-Redundancy-Maximal-Relevance (mRMR) method is a very famous feature selection algorithm that uses mutual information $I(F_i;C)$ to measure class-relevance, i.e.
$$I(F_i;C)$$
where $F_i$ is the candidate feature, and $I(F_i;F_s)$ to measure paired dependency, i.e.
$$\frac{1}{|\s{S}|}\sum_{F_s\in\s{S}}I(F_i;F_s)$$
where $\s{S}$ is the already-selected feature subset. It selects feature satisfying 
$$\max_{F_i}\left\{I(F_i;C)-\frac{1}{|\s{S}|}\sum_{F_s\in\s{S}}I(F_i;F_s)\right\}$$
in a greedy manner. It is a kind of feature selection methods that only consider pairwise redundancy \cite{FEAST}.
\item DISR algorithm \cite{DISR}: The Double Input Symmetrical Relevance method uses following criterion to selected features:
$$J_{disr}(F_i)=\sum_{F_s\in\s{S}}\frac{I(F_iF_s;C)}{H(F_iF_sC)}$$
where $H(\cdot)$ denotes information entropy. Features with maximum $J_{disr}$ will be selected by DISR. This criterion is a modification of the Joint Mutual Information criterion \cite{JMI} which is shown to be one of the most effective MI-based feature selection criteria as indicated in \cite{FEAST}. Also, It is a kind of feature selection methods that only consider pairwise redundancy \cite{FEAST}.
\item MIM algorithm \cite{MIM}: The Mutual Information Maximization algorithm harnesses a simple criterion that only measure the class-relevance of the candidate features and select feature satisfying
$$\max_{F_i} I(F_i;C)$$
in a greedy manner. This heuristic, which considers a score for each feature independently of others, has been used many times in the literature. Note that it is also called Information Gain criterion sometimes because they share the same formulation in feature selection.
\item ReliefF algorithm \cite{ReliefF}: The ReliefF algorithm is an extension of Relief algorithm \cite{Relief}. It is a well-known distance-based feature selection method that searches nearest neighbors of samples for each class label and then weights features in terms of how well they differentiate samples for different class labels. For ReliefF, we use 5 neighbors and 30 instances as suggested by \cite{ReliefF} throughout the experiments.
\item DEAFS \cite{DEAFS}: DEAFS is the first feature selection algorithm based on Data Envelopment Analysis. Like most of the typical filters, it evaluates according to more than one criterion (i.e. relevance, redundancy, conditional dependence, etc.), but does not do simple arithmetic operations like what mRMR and DISR do as introduced above. Instead, it applies DEA-based evaluation approach to get the overall score of a feature by considering its relevance to class and conditional dependence to every other feature in the feature space. Moreover, it is a one-time ranking algorithm (like MIM) rather than iterating so many times (like mRMR and DISR) to get final feature ranking results.
\end{itemize}

We conduct experiments for proposed as well as selected feature selection methods by comparing the classification results on the datasets constructed with the top $1,...,m$ features. The upper bound of $m$ in our experiment is $\min\{|\s{F}|,30\}$. Since joint conditional mutual information estimation may result in a premature end of the selecting process, we set the upper bound of $m$ to be $\min\{|\s{S}|,30\}$, where $\s{S}$ is the set of actually-selected features of DEA-CS. To achieve impartial results, we conduct 10-fold cross-validation and report the average results of four classifiers. Since DEA-CS and DEAFS are implemented in Java and MATLAB (for LP solving), while mRMR, DISR, and MIM are called in FEAST implemented in C++ \cite{FEAST} and ReliefF is called in Weka implemented in Java, runtime comparison for these feature selection methods makes no sense and thus it will not be conducted and reported in our experiments. In fact, from the time complexity analysis we can find that proposed DEA-CS is with similar time complexity level like typical incremental feature ranking algorithms such as mRMR and DISR. However, since MATLAB code runs much slower than C++, DEA-CS is more time-consuming than MIM, ReliefF, and even mRMR and DISR. We have to admit that this is a drawback of our work and we try to use much more efficient LP solver such as gurobi and CPLEX to significantly promote the execution efficiency in our future work. All experiments are conducted on a personal computer with Windows 7, 3.40 GHz quad-core CPU, and 8GB memory.

\subsection{Classification results and discussion}
Figs. \ref{flags}--\ref{gisette} represent the average classification accuracies of four classifiers on ten datasets to illustrate the effect of changes of the performance with respect to an incremental scale of selected features, where the X axis depicts the consecutive numbers of selected features, and Y axis depicts the average classification accuracy. The best performance of each algorithm on the selected datasets is also reported in Tab. \ref{best_res}.

\begin{figure*}[!ht]
\begin{center}
  \includegraphics[scale = 0.62]{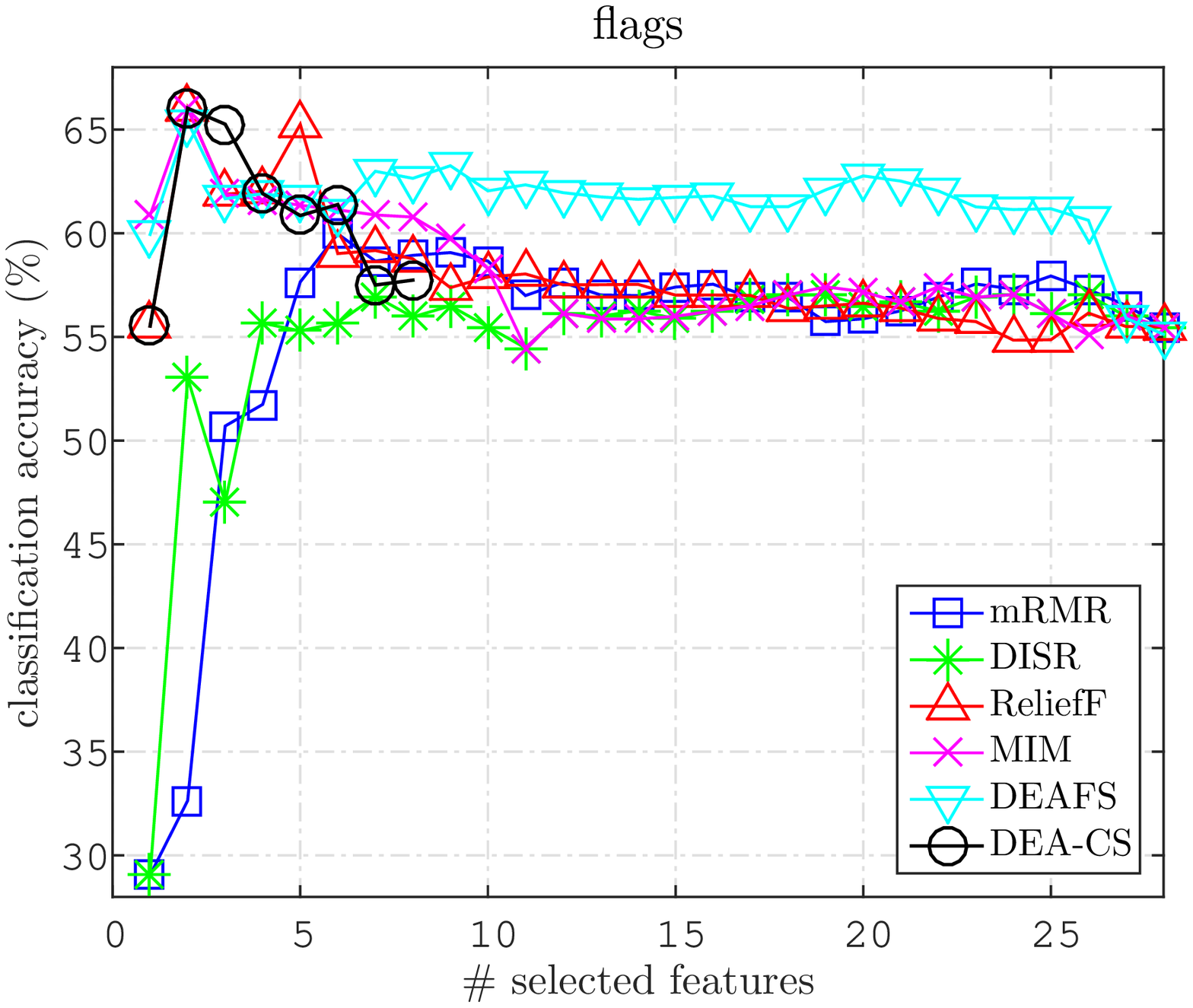}\\
  \caption{Accuracy comparison with different number of selected features on flags dataset.}\label{flags}
\end{center}
\end{figure*}

\begin{figure*}[!ht]
\begin{center}
  \includegraphics[ scale = 0.62]{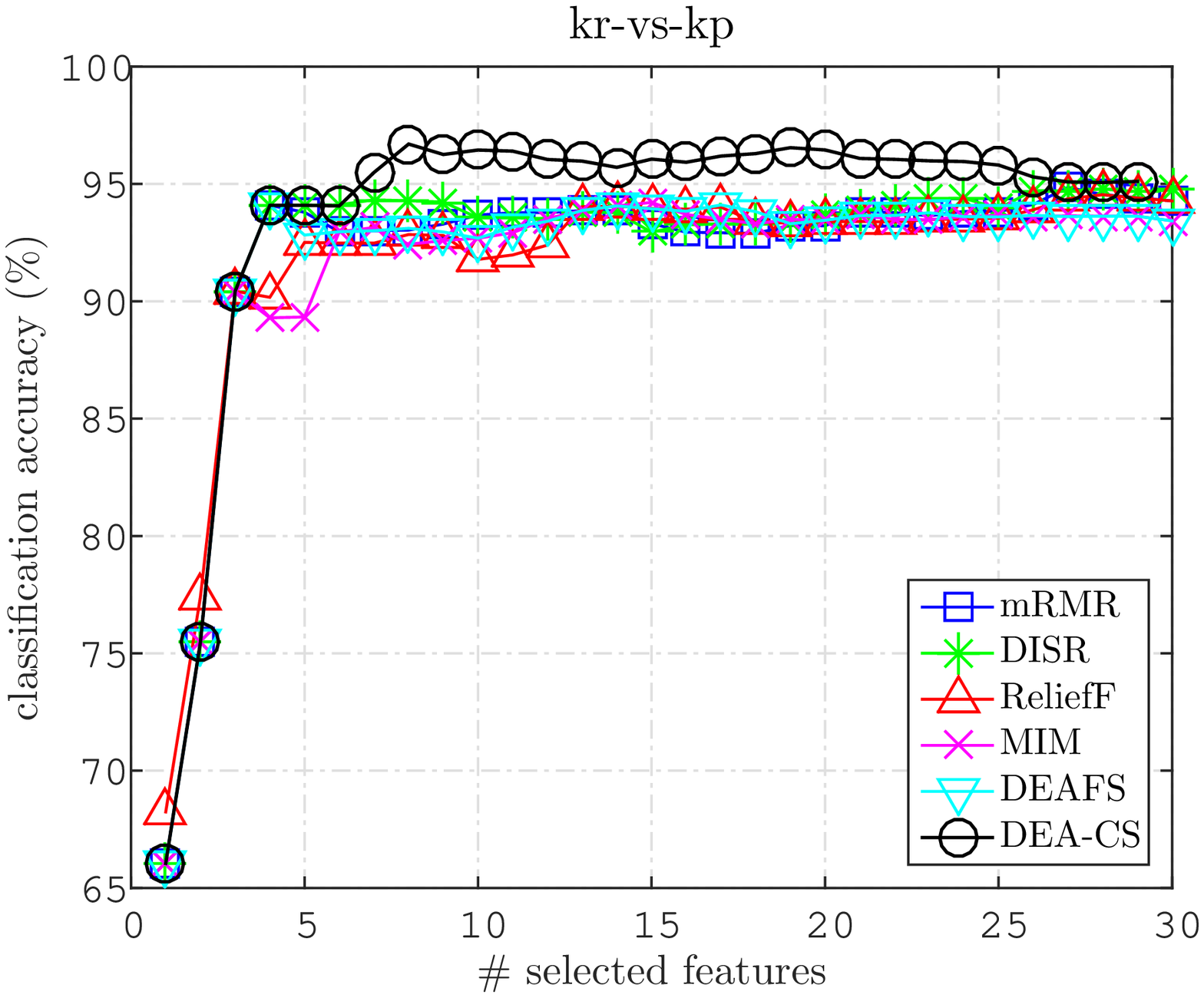}\\
  \caption{Accuracy comparison with different number of selected features on kr-vs-kp dataset.}\label{kr_vs_kp}
\end{center}
\end{figure*}

\begin{figure*}[!ht]
\begin{center}
  \includegraphics[ scale = 0.62]{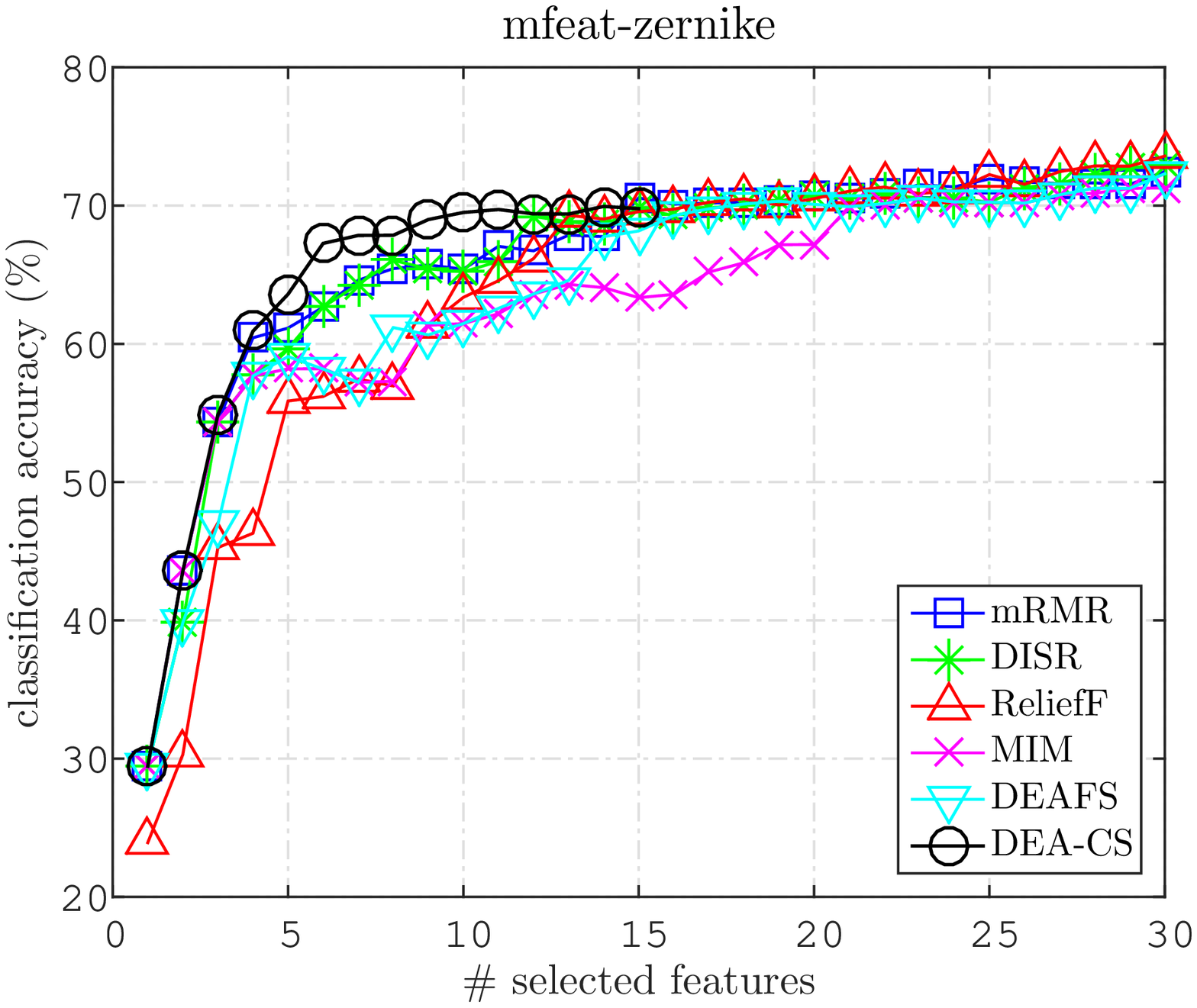}\\
  \caption{Accuracy comparison with different number of selected features on mfeat-zernike dataset.}\label{zernike}
\end{center}
\end{figure*}

\begin{figure*}[!ht]
\begin{center}
  \includegraphics[ scale = 0.62]{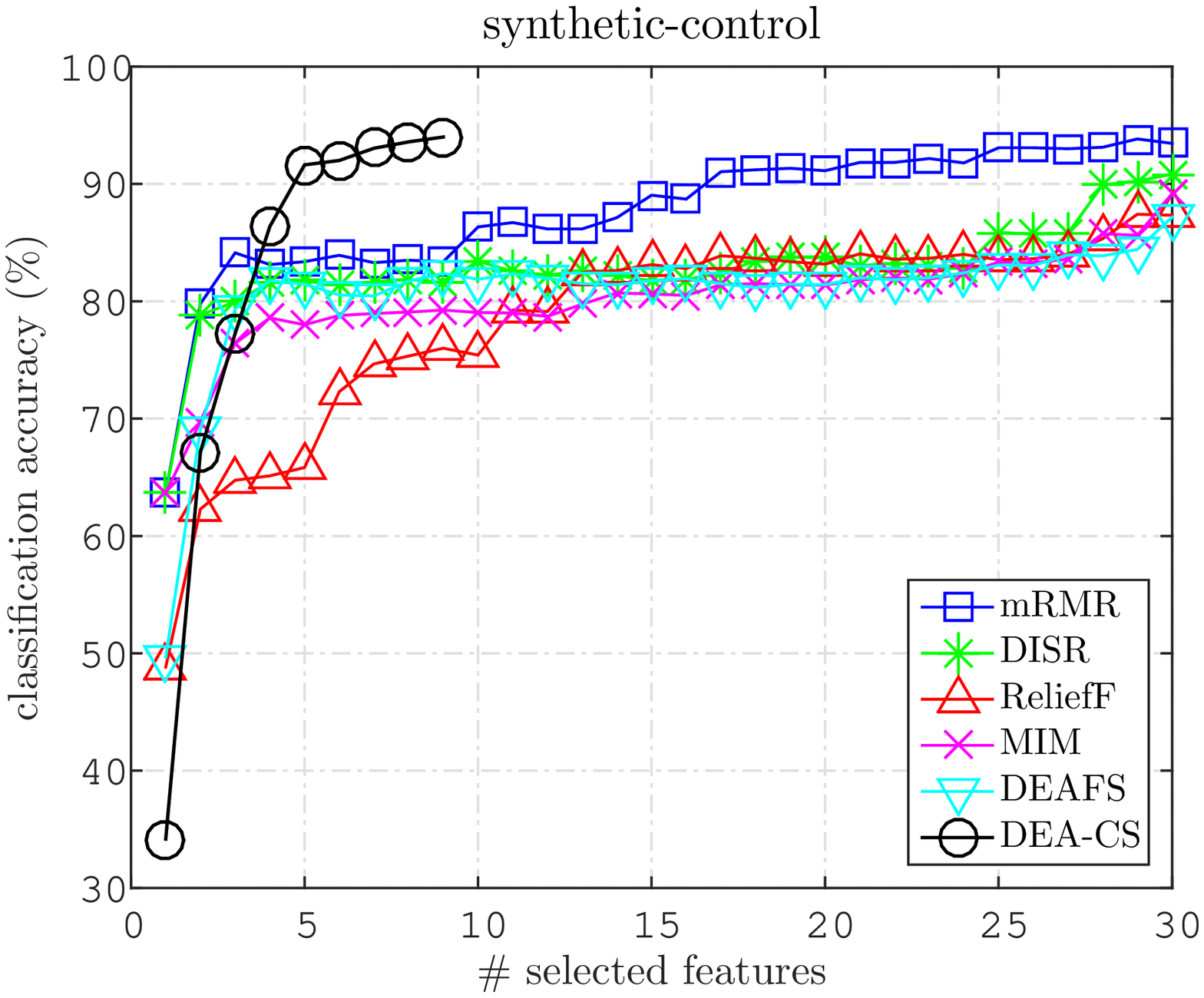}\\
  \caption{Accuracy comparison with different number of selected features on synthetic-control dataset.}\label{synthetic}
\end{center}
\end{figure*}

\begin{figure*}[!ht]
\begin{center}
  \includegraphics[ scale = 0.62]{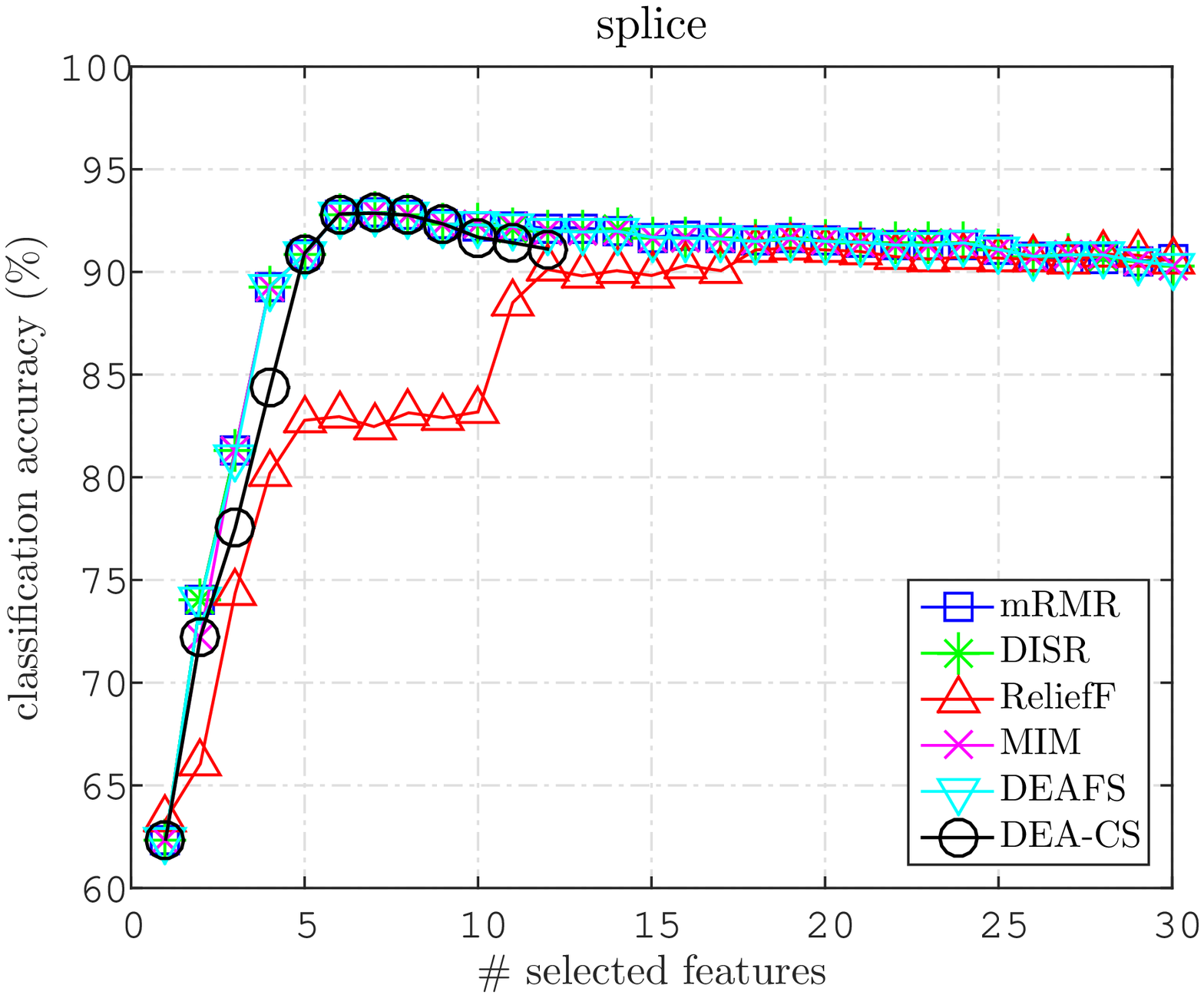}\\
  \caption{Accuracy comparison with different number of selected features on splice dataset.}\label{splice}
\end{center}
\end{figure*}

\begin{figure*}[!ht]
\begin{center}
  \includegraphics[ scale = 0.62]{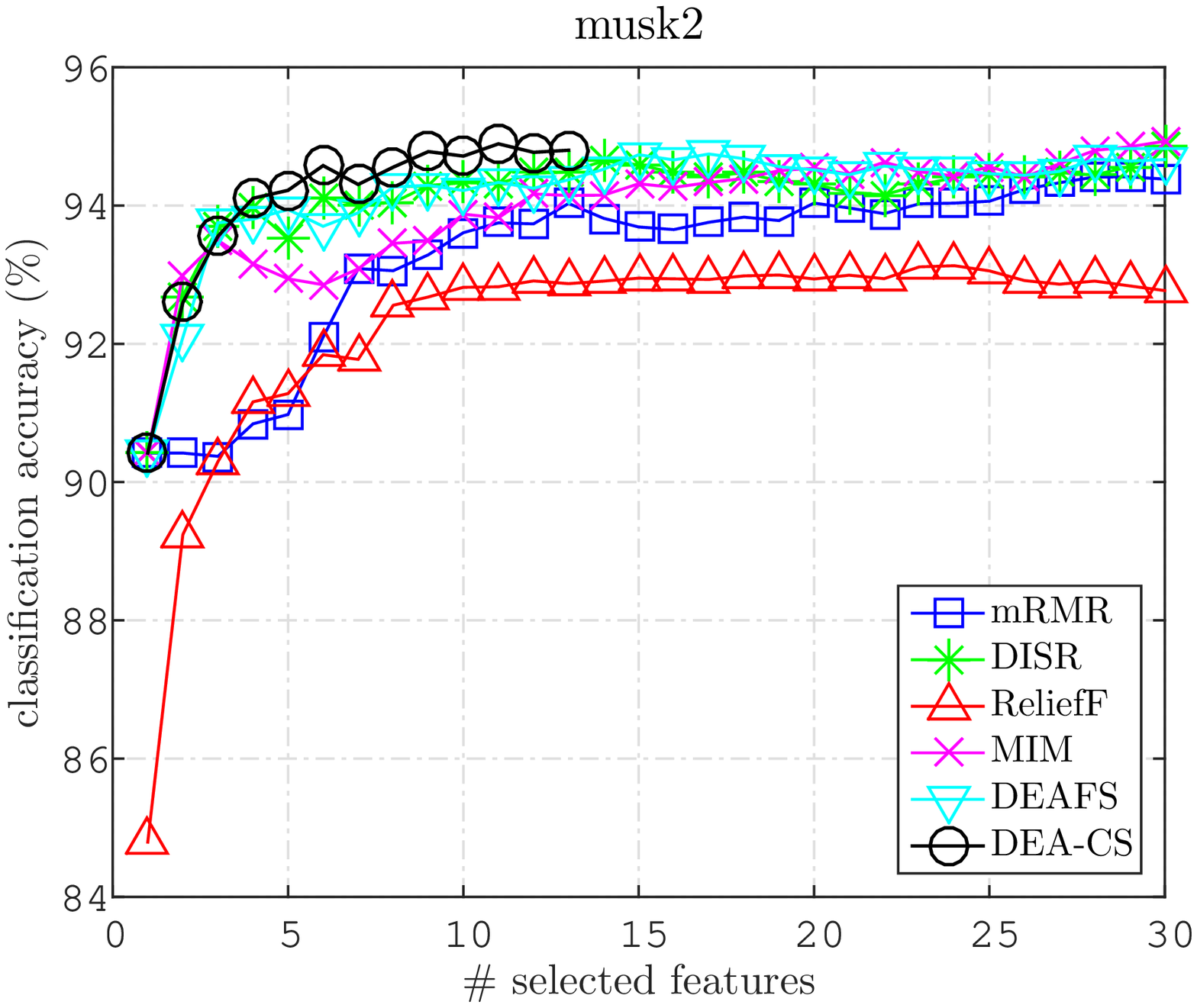}\\
  \caption{Accuracy comparison with different number of selected features on musk2 dataset.}\label{musk2}
\end{center}
\end{figure*}

\begin{figure*}[!ht]
\begin{center}
  \includegraphics[scale = 0.62]{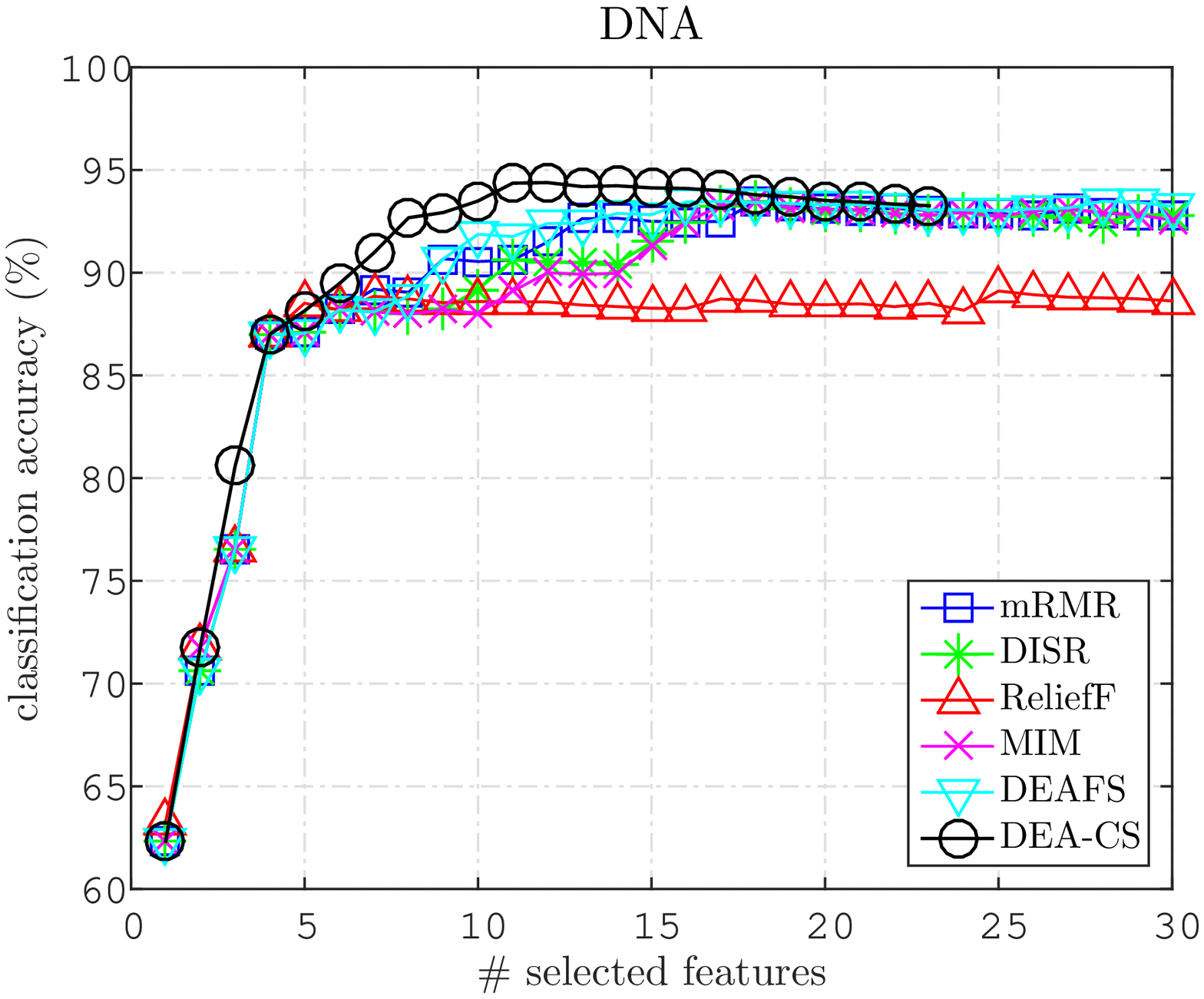}\\
  \caption{Accuracy comparison with different number of selected features on DNA dataset.}\label{DNA}
\end{center}
\end{figure*}

\begin{figure*}[!ht]
\begin{center}
  \includegraphics[scale = 0.62]{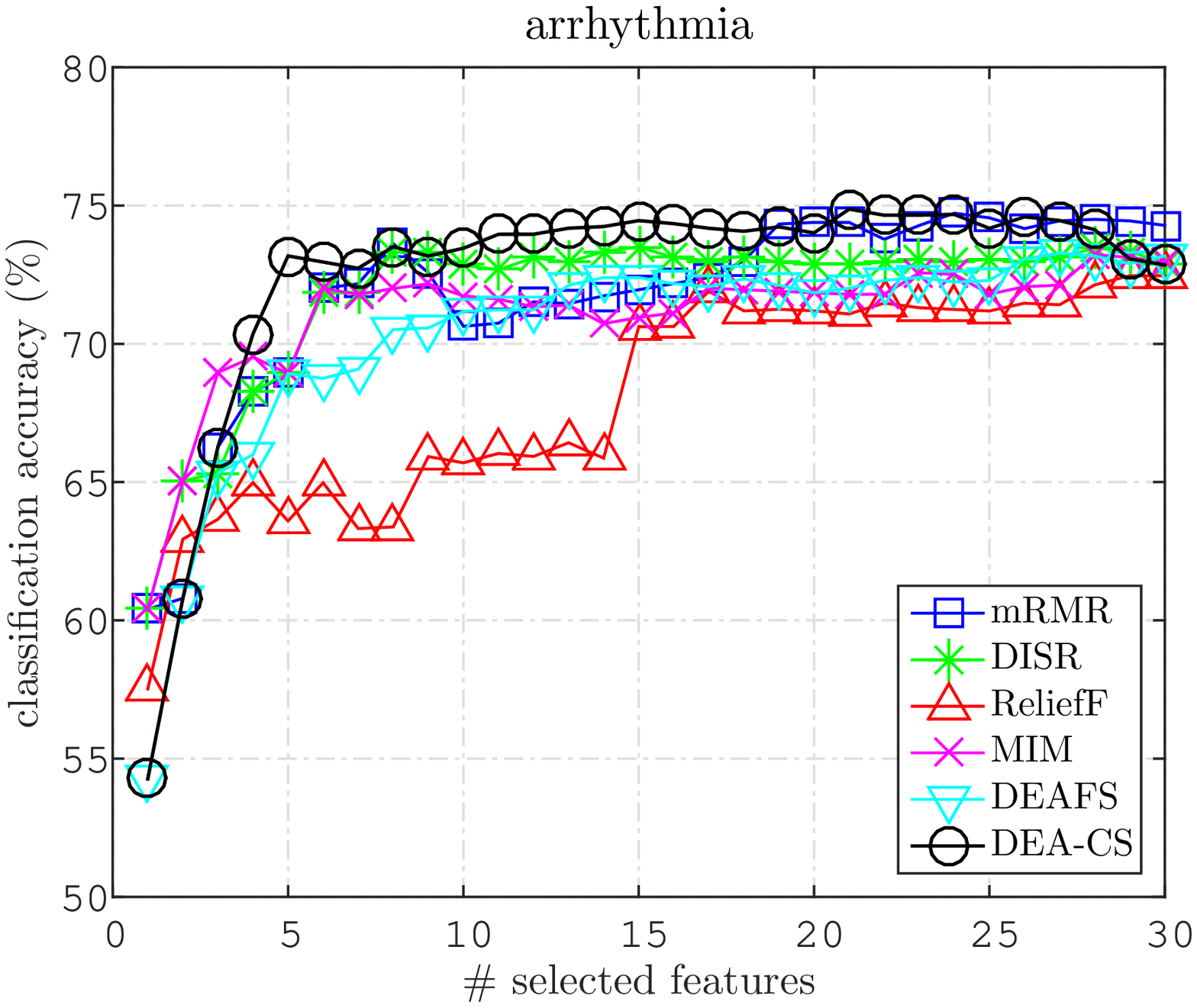}\\
  \caption{Accuracy comparison with different number of selected features on arrhythmia dataset.}\label{arrhythmia}
\end{center}
\end{figure*}

\begin{figure*}[!ht]
\begin{center}
  \includegraphics[scale = 0.62]{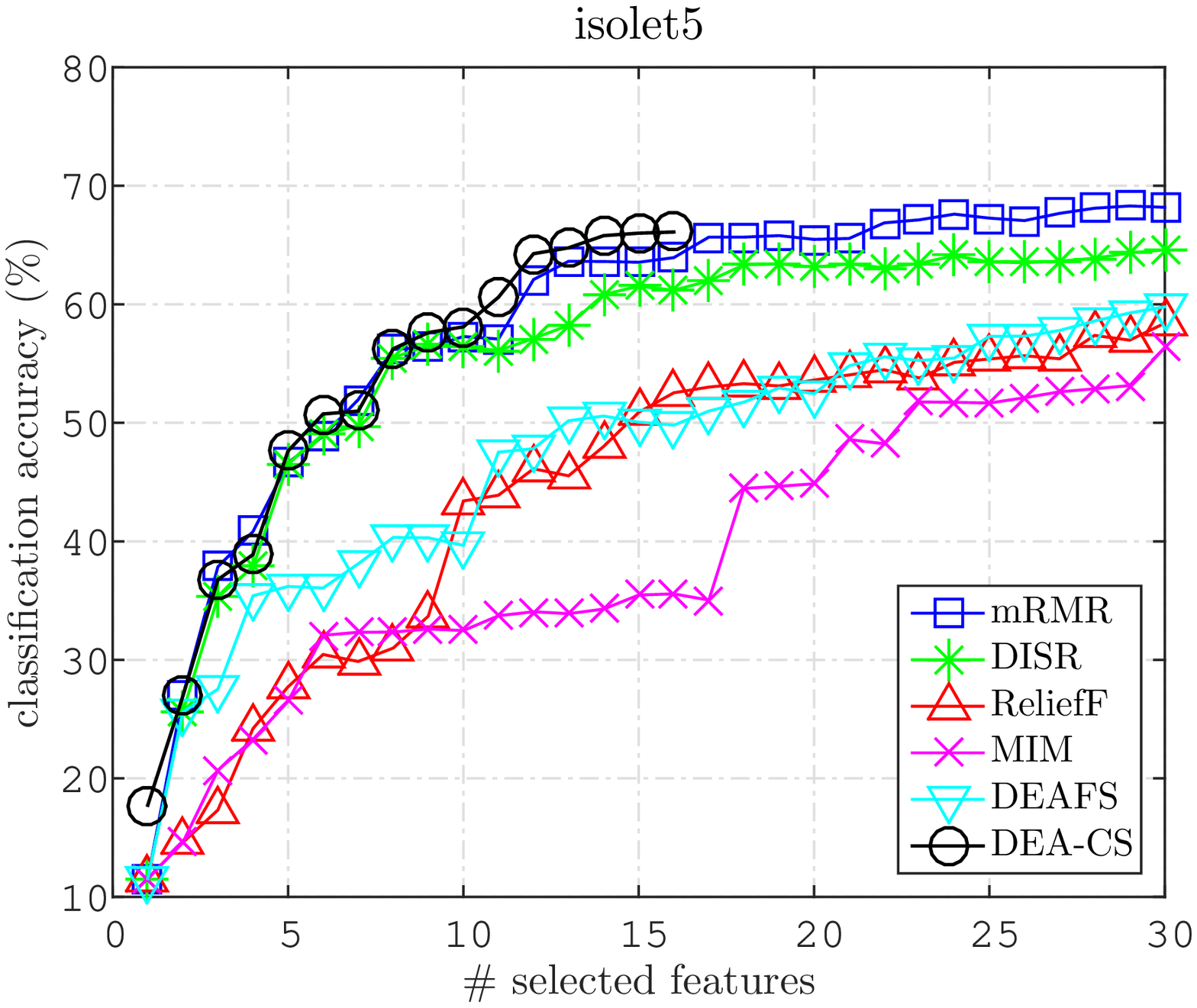}\\
  \caption{Accuracy comparison with different number of selected features on isolet5 dataset.}\label{isolet5}
\end{center}
\end{figure*}

\begin{figure*}[!ht]
\begin{center}
  \includegraphics[scale = 0.62]{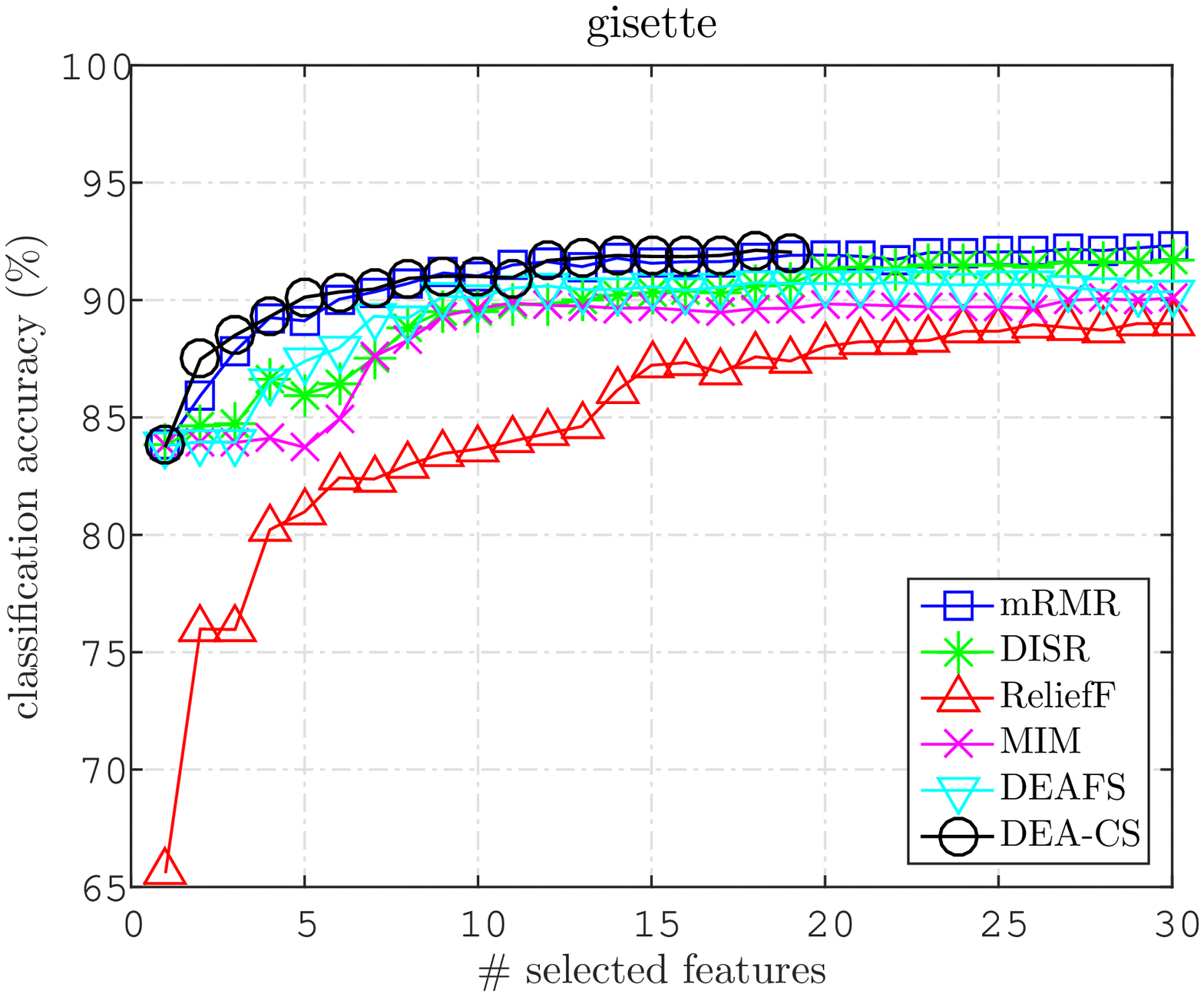}\\
  \caption{Accuracy comparison with different number of selected features on gisette dataset.}\label{gisette}
\end{center}
\end{figure*}

\begin{table*}[!ht]
	\begin{threeparttable}
		\centering
		\caption{Average accuracy of NBC, SVM, $k$NN, and C4.5 on the selected features: Columns \% and \# record the best average accuracy rate and the number of the selected features corresponding to the best accuracy for the current feature selection algorithm, respectively. Recall that the upper bound of selected features in our experiment is $\min\{|F|, 30\}$.}
		\begin{tabular}{r l r p{0.2cm} @{\hspace{3pt}}c r p{0.2cm} @{\hspace{3pt}}c r p{0.2cm} @{\hspace{3pt}}c r p{0.2cm} @{\hspace{3pt}}c r p{0.2cm} @{\hspace{3pt}}c r p{0.2cm} @{\hspace{3pt}}c}
			\toprule
			\multirow{2}{30pt}{Dat. No.}& \multicolumn{2}{c}{DEA-CS}&&\multicolumn{2}{c}{DEAFS}&  &\multicolumn{2}{c}{mRMR}&  &\multicolumn{2}{c}{DISR}&  &\multicolumn{2}{c}{ReliefF}&  &\multicolumn{2}{c}{MIM}&  \\
			\cline{2-3}\cline{5-6}\cline{8-9}\cline{11-12}\cline{14-15}\cline{17-18}
			&  \#   & \%\phantom{a} & & \# & \% \phantom{a}&  & \#  & \%\phantom{a}&  & \#   & \%\phantom{a}&  & \#   & \%\phantom{a}&  & \#   & \%\phantom{a}&  \\
			\midrule
			1\phantom{000} & 2  & \bf{66.03}  &  & 2 & 65.38     &  & 6  & 59.97    &  & 19& 57.04      &  & 2  &\bf{66.03}&  & 2 &\bf{66.03} &  \\
			2\phantom{000} & 8  & \bf{96.70}  &  & 17& 94.11     &  & 27 & 94.85    &  & 28& 94.78      &  & 28 & 94.54    &  & 15& 94.21     &  \\
			3\phantom{000} & 14 &     69.94   &  & 30& 72.30     &  & 30 & 72.45    &  & 30& 72.94      &  & 30 &\bf{73.58}&  & 30& 71.28     &  \\
			4\phantom{000} & 9  & \bf{94.00}  &  & 30& 87.17     &  & 29 & 93.84    &  & 30& 90.83      &  & 29 & 87.42    &  & 30& 89.17     &  \\
			5\phantom{000} & 7  & \bf{92.89}  &  & 7 &\bf{92.89} &  & 7  &\bf{92.89}&  & 7 &\bf{92.89}  &  & 19 & 91.16    &  & 7 & \bf{92.89}&  \\
			6\phantom{000} & 11 & \bf{94.89}  &  & 17& 94.75     &  & 29 & 94.42    &  & 30& 94.85      &  & 24 & 93.13    &  & 30& 94.85     &    \\
			7\phantom{000} & 12 & \bf{95.41}  &  & 28& 93.49     &  & 18 & 93.48    &  & 18& 93.48      &  & 25 & 89.11    &  & 18& 93.48     &    \\
			8\phantom{000} & 21 & \bf{74.87}  &  & 27& 73.29     &  & 24 & 74.72    &  & 15& 73.51      &  & 29 & 72.46    &  & 28& 73.23     &    \\
			9\phantom{000} & 16 & 66.10       &  & 30& 59.79     &  & 29 &\bf{68.30}&  & 30& 64.60      &  & 30 & 58.41    &  & 30& 56.35     &   \\
			10\phantom{000}& 18 & 92.12       &  & 22& 90.76     &  & 30 &\bf{92.32}&  & 30& 91.65      &  & 29 & 89.00    &  & 30& 90.09     &  \\
			&&&&&&&&&&&&&&&&&&\\
			Avg.\phantom{00} &11.8&\bf{84.19} &  & 21.0& 82.39   &  & 22.9& 83.72   &  &23.7& 82.65     &  & 24.5 & 81.48  &  & 22.0& 82.16   &  \\
			\bottomrule
		\end{tabular}%
		\label{best_res}%
	\end{threeparttable}%
\end{table*}

Results shown in Figs. \ref{flags}--\ref{gisette} and Tab. \ref{best_res} taht asdf have verified the superiority of DEA-CS according to the classification results. DEA-CS performs better in most of the cases, particularly on five datasets namely flags (Fig. \ref{flags}), kr-vs-kp (Fig. \ref{kr_vs_kp}), synthetic-control (Fig. \ref{synthetic}), musk2 (Fig. \ref{musk2}), and DNA (Fig. \ref{DNA}), and is comparable to other feature selection algorithms on rest of the datasets. According to the results shown in Tab. \ref{best_res}, DEA-CS holds the best average accuracy among six feature selection methods 7 times, and also wins the best average accuracy $84.89\%$ on all selected datasets. Furthermore, the number of its average best selected features is only 11.8, whereas the second minimum number of the average best selected features (held by DEAFS) is 21.0. This illustrates that DEA-CS performs extraordinarily better than other methods when the size of selected features is small, and also verifies the superiority of the estimation of $Div(F;c_i|\s{S})$ conditioned on the whole subset $\s{S}$ comparing to the pairwise approximation applied by mRMR and DISR. More specifically, DEA-CS usually performs mediocrely at the very beginning of the top-selected features, but superiorly after the coming of the following ones (which can be found clearly on synthetic-control (Fig. \ref{synthetic}) and DNA (Fig. \ref{DNA})). This is probably because taking all of the selected features as the conditioning set when measuring conditional independence of features, rather than only measuring pairwise redundancy like mRMR, DISR, and DEAFS, can effectively avoid local optima.  We also find that ReliefF as well as MIM performs worst among the four selectors in most cases. This is partially due to their ignorance of redundancy among features. For ReliefF which performs worst, it may also lie in no superior parameters preselected to cater various kinds of datasets. DISR, mRMR, and DEAFS perform better than ReliefF and MIM since they consider feature redundancy. However, for DISR and mRMR in some cases, e.g. on flags and kr-vs-kp, they seem to be not so effective. The possible reason is that they only consider pairwise rather than $k$-wise ($k\ge 3$) redundancy among features. For mRMR, it may also lie in the shortages of unsupervised redundancy analysis as indicated in \cite{DEAFS}. For DEAFS, it performs well on flags, splice, DNA, and musk2, which indicate the effectiveness of DEA-based feature evaluation strategy. However, it performs not so well on mfeat-zernike and synthetic-control, and is even inferior to most of the other methods on isolet5. This possibly lies in that there are too many outputs (i.e. conditional dependence to every other feature) for the DEA-based evaluation process, which breaks rule-of-thumb in DEA \cite{rule_of_thumb_DEA} suggesting that the number of DMUs should be at least three times the total number of inputs plus outputs used in the models. In addition, due to considering relationship to every other feature in the feature space, the conditional dependence between a current candidate feature and other actually-useless features may obstacle the evaluation process of this feature and thus finally impairs the quality of selected features. As for proposed method DEA-CS, the number of outputs (determined by the number of class labels) is usually far smaller than the sample size and thus making DEA-CS never break the rule-of-thumb in DEA in most cases. In addition, DEA-CS measures conditional dependence of the candidate feature to only the current selected features rather than to all other features in the feature space consisting of both actually-useful and actually-useless ones, thus making the evaluation process of DEA-CS more effective than that of DEAFS.

We may also find that performance of DEA-CS is not outstanding and sometimes inferior to mRMR and DISR. In addition, DEA-CS selects less features than other methods (e.g. it can only select at most eight and nine features on flags and synthetic-control, respectively). This implies that DEA-CS ends its selecting iteration because of $R(F;c_i|\s{S})=0$ for all $F$ and $c_i$ (see the 16th step in Algorithm \ref{alg}). This is due to the sample inefficiency when estimating joint probability on $\s{S}$ consisting more than medium amount of features (e.g. more than 10 features), which finally results in inaccurate estimate values that will obstacle the selection process. That is, highly redundant features may be selected and they significantly impair the classification accuracy. It can also illustrate why DEA-CS often performs inferiorly when the scale of selected feature grows at the late stage. In addition, the super-efficiency DEA formula applied in DEA-CS evaluates features in a Debreu-Farrell manner, which is input-oriented and neglects slacks for both inputs and outputs. It is a kind of radial measures and features may be inaccurately evaluated for it never considers trade-offs between inputs or between outputs. To this end, alternative non-radial DEA that estimate technical inefficiency with Pareto-Koopmans measures \cite{Enhanced_Russell,RAM,BAM} could be considered for performance promotion.

To further illustrate the feature selection process of DEA-CS, we give an example of the selecting process at the $13$th iteration of DEA-CS on gisette dataset (Fig. \ref{example_gisette}) which only contains two class labels ($C=\{c_1,c_2\}$) and thus fits for 2-D illustration.

\begin{figure*}[!ht]
\begin{center}
  \includegraphics[trim = 7.5em 1em 0 0, scale = 0.6]{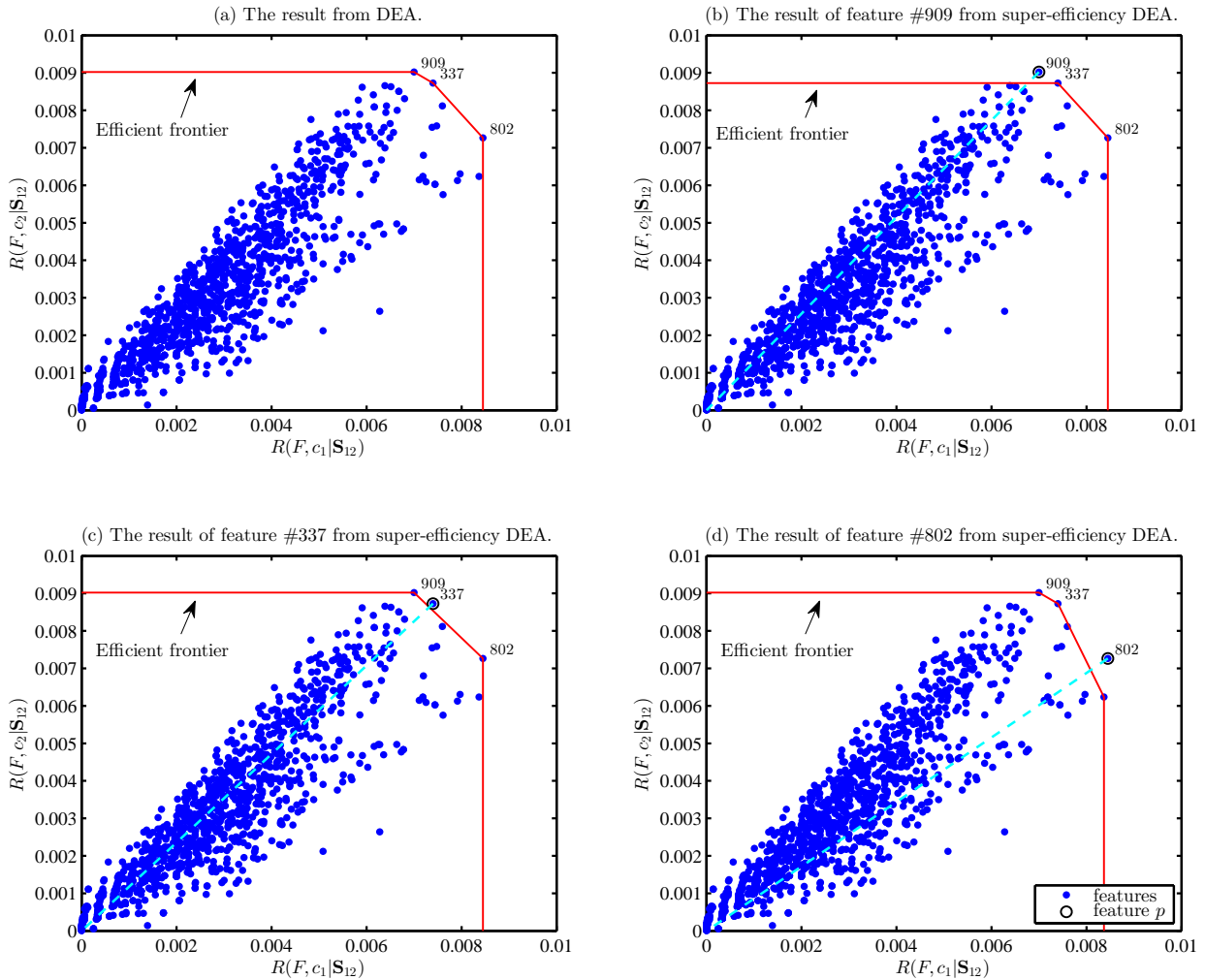}\\
  \caption{An example of the selecting process on gisette dataset.}\label{example_gisette}
\end{center}
\end{figure*}

We can see that, in Fig. \ref{example_gisette}, twelve features have already been selected in $\s{S}_{12}$ and DEA-CS is currently searching for the $13$th feature, where X and Y axes depict the value of $R(F;c_1|\s{S}_{12})$ and $R(F;c_2|\s{S}_{12})$ respectively. From Fig. \ref{example_gisette} (a) we can see that $F_{337},F_{802}$, and $F_{909}$ have been evaluated as efficient features and thus locate on the efficient frontier (the red line), hence the selected features will be determined among them in terms of their super-efficiency scores. Fig. \ref{example_gisette} (b) -- (d) show the new frontiers correspond to $F_{909}$, $F_{337}$, and $F_{802}$, respectively, where each of them is constructed by the candidate features without the corresponding feature (we use feature $p$ in Fig. \ref{example_gisette} (b) -- (d) to denote the currently-evaluated feature (i.e. the corresponding feature)). It is obvious that $F_{802}$ is furthest from the super-efficient frontier, which implies its score will be larger than other two features, and thus it will be selected by DEA-CS. 
However, if simply adding $R(F;c_1|\s{S}_{12})$ and $R(F;c_2|\s{S}_{12})$ together, like what most MI-based feature selection methods do (which is mentioned in section \ref{CS-strategy}), $F_{909}$ would get the largest score and finally be selected \footnote{Note that sum of $R(F;c_1|\s{S}_{12})$ and $R(F;c_2|\s{S}_{12})$ does not equal to $I(F;C|\s{S}_{12})$, and thus is not applied by any MI-based feature selection methods.}. From Fig. \ref{gisette}, we can see that DEA-CS outperforms other feature selection methods when all methods select 13 features.

\section{Conclusions \& future work}\label{conclusion}

In this paper, we take feature selection as a multi-index evaluation process and propose a novel feature selection method based on CS strategy and super-efficiency DEA. Since salient features have strong discriminative power and it would not be impaired given other features, conditional dependence between the feature-under-evaluation and class given the currently-selected subset is applied to conduct relevance and redundancy analysis. Different from the extant MI-based feature selection methods, each class label is taken as an individual variable and CS strategy is then applied in our method to explicitly handle relevance and redundancy for them. In order to compare the conditional dependence distributions of the candidate features, each candidate is taken as a DMU with $|C|$ outputs (i.e. $|C|$ conditional dependence scores) and a constant input, and a super-efficiency DEA is employed to evaluate the candidate features. The feature with maximum efficiency score will be finally selected in the feature subset which will be used as the conditioning subset in the next iteration, in such a way as to make the scale of the currently-selected subset increased incrementally. To validate the effectiveness of DEA-CS, we compare it with four representative and widely-used feature selection methods namely mRMR, DISR, MIM, and ReliefF through several classification experiments on ten UCI datasets. Empirical results show that proposed method outperforms other feature selection methods in most cases.

However, DEA-CS encounters the difficulty of sample inefficiency for joint distribution estimation, since it always uses the currently-selected subset achieved at the last iteration as the conditioning variable when estimating conditional dependence of candidate features. This is a fatal drawback of the estimation approach. Although the CS strategy applied in DEA-CS estimates $R(F;c|\s{S})$ on the whole samples and thus can partially mitigate the negative effects caused by sample inefficiency, exponential growth of the sample requirement caused by a steady increase of the scale of the feature subset cannot be avoided in many real world datasets, particularly in bioinformatic datasets which contain very few samples with thousands of features. Therefore, improved pairwise-approximation criterion will be taken into consideration in future work to tackle this problem.

In addition, as we mentioned before, the super-efficiency DEA applied in DEA-CS is a radical DEA approach, which evaluates DMUs in a Debreu-Farrell input-oriented manner and thus neglects slacks for both inputs and outputs. Features may thus be inaccurately evaluated for it never considers trade-offs between inputs or between outputs. Alternative Pareto-Koopmans DEA measures, e.g. the Enhanced Russell Graph \cite{Enhanced_Russell}, RAM \cite{RAM}, and BAM \cite{BAM}, are possible to be considered in future to make trade-offs between inputs or between outputs.

\section*{Acknowledgement}
The authors would like to thank the anonymous referees for their constructive comments which were very helpful in revising this paper. This work is partially supported by National Natural Foundation of China (71320107001), the Fundamental Research Funds for the Central Universities, HUST (CXY12Q044, CXY13Q035), the Graduates' Innovation Fund of Huazhong University of Science \& Technology (No. HF-11-20-2013), and the Doctorate Fellowship Foundation of Huazhong University of Science \& Technology (D201177780). The first and fifth authors acknowledge the financial support from China Scholarship Council (CSC).

\section*{References}
\bibliographystyle{elsarticle-num}
\hspace*{\stretch{1}}
\bibliography{bibdata}

\begin{thebibliography}{10}
\expandafter\ifx\csname url\endcsname\relax
  \def\url#1{\texttt{#1}}\fi
\expandafter\ifx\csname urlprefix\endcsname\relax\def\urlprefix{URL }\fi
\expandafter\ifx\csname href\endcsname\relax
  \def\href#1#2{#2} \def\path#1{#1}\fi

\bibitem{InfoSourceSel}
Y.~Lin, X.~Hu, X.~Wu, Quality of information-based source assessment and
  selection, Neurocomputing 133 (2014) 95--102.

\bibitem{MI_DSS}
S.~Cang, H.~Yu, Mutual information based input feature selection for
  classification problems, Decision Support Systems 54 (2012) 691--698.

\bibitem{2}
I.~Guyon, A.~Elisseeff, An introduction to variable and features election,
  Journal of Machine Learning Research 3 (2003) 1157--1182.

\bibitem{4}
H.~Liu, L.~Yu, Toward integrating features election algorithms for
  classification and clustering, IEEE Transactions on Knowledge and Data
  Engineering 17~(4) (2005) 491--502.

\bibitem{6}
T.~S. Furey, N.~Cristianini, N.~Duffy, D.~W. Bednarski, M.~Schummer,
  D.~Haussler, Support vector machine classification and validation of cancer
  tissue samples using microarray expression data, Bioinformatics 16~(10)
  (2000) 906--914.

\bibitem{8}
G.~Qu, S.~Hariri, M.~Yousif, A new dependency and correlation analysis for
  features, IEEE Transactions on Knowledge and Data Engineering 17~(9) (2005)
  1199--1207.

\bibitem{9}
H.~Peng, F.~Long, C.~Ding, Feature selection based on mutual information:
  criteria of max-dependency, max-relevance, and min-redundancy, IEEE
  Transactions on Pattern Analysis and Machine Intelligence 27~(8) (2005)
  1226--1238.

\bibitem{11}
D.~Huang, T.~W.~S. Chow, Effective feature selection scheme using mutual
  information, Neurocomputing 63 (2005) 325--343.

\bibitem{12}
J.~J. Huang, Y.~Z. Cai, X.~M. Xu, A parameterless feature ranking algorithm
  based on mi, Neurocomputing 71 (2008) 1656--1668.

\bibitem{13}
L.~Yu, H.~Liu, Efficient feature selection via analysis of relevance and
  redundancy, Journal of Machine Learning Research 5 (2004) 1205--1224.

\bibitem{Le_Song}
L.~Song, A.~Smola, A.~Gretton, J.~Bedo, K.~Borgwardt, Feature selection via
  dependence maximization, Journal of Machine Learning Research 13 (2012)
  1393--1434.

\bibitem{Limit_MI}
B.~Fr\'{e}nay, G.~Doquire, M.~Verleysen, Theoretical and empirical study on the
  potential inadequacy of mutual information for feature selection in
  classification, Neurocomputing 112 (2013) 64--78.

\bibitem{14}
Z.~Zhao, H.~Liu, Spectral feature selection for supervised and unsupervised
  learning, in: Proceedings of the 24th International Conference on Machine
  learning, ICML'07, ACM Press, New York, NY, USA, 2007, pp. 1151--1157.

\bibitem{15}
L.~Song, A.~J. Smola, K.~M. Borgwardt, J.~Bedo, Supervised feature selection
  via dependence estimation, in: Proceedings of the 24th International
  Conference on Machine learning, ICML'07, ACM Press, New York, NY, USA, 2007,
  pp. 823--830.

\bibitem{CFS}
M.~A. Hall, Correlation-based feature selection for discrete and numeric class
  machine learning, in: Proceedings of the 7th International Conference on
  Machine Learning, ICML'00, Morgan Kaufmann, Los Altos, CA, USA, 2000, pp.
  359--366.

\bibitem{NPHard}
E.~Amaldi, V.~Kann, On the approximation of minimizing non zero variables or
  unsatisfied relations in linear systems, Theoretical Computer Science 209
  (1998) 237--260.

\bibitem{NPComplete}
A.~A. Albrecht, Stochastic local search for the feature set problem, with
  applications to microarray data, Applied Mathematics and Computation 183~(2)
  (2006) 1148--1164.

\bibitem{MIFS}
R.~Battiti, Using mi for selecting features in supervised neural net learning,
  IEEE Transactions on Neural Networks 5~(4) (1994) 537--550.

\bibitem{MIFS-U}
N.~Kwak, C.~H. Choi, Input feature selection for classification problems, IEEE
  Transactions on Neural Networks 13~(1) (2002) 143--159.

\bibitem{CMIM}
F.~Fleuret, Fast binary feature selection with conditional mutual information,
  Journal of Machine Learning Research 5 (2004) 1531--1555.

\bibitem{Relief}
K.~Kira, L.~Rendell, Apractical approach to feature selection, in:
  Proceedingsof the 9th International Workshop on Machine Learning, ML'92,
  Morgan Kaufmann, San Francisco, CA, USA, 1992, pp. 249--256.

\bibitem{Var_Of_relief}
I.~Kononenko, Estimating attributes: analysis and extensions of relief, in:
  Proceedings of European Conference on Machine Learning, ECML'94,
  Springer-Verlag New York, Inc., Secaucus, NJ, USA, 1994, pp. 171--182.

\bibitem{MIM}
D.~D. Lewis, Feature selection and feature extraction for text categorization,
  in: Proceedings of the workshop on Speech and Natural Language, Association
  for Computational Linguistics Morristown, NJ, USA, 1992, pp. 212--217.

\bibitem{DEAFS}
Y.~Zhang, A.~Yang, C.~Xiong, Z.~Zhang, Feature selection using data envelopment
  analysis, Knowledge-Based Systems 64 (2014) 70--80.

\bibitem{mRMR30}
A.~R. Webb, Statistical Pattern Recognition, 2nd Edition, John Wiley \& Sons
  Ltd, Chichester, West Sussex, England, 2002.

\bibitem{KollerSahami}
D.~Koller, M.~Sahami, Toward optimal feature selection, in: Proceedings of
  International Conference on Machine Learning, ICML'96, Morgan Kaufmann, Los
  Altos, CA, USA, 1996, pp. 284--292.

\bibitem{Kohavi_John}
R.~Kohavi, G.~H. John, Wrappers for feature subset selection, Artificial
  Intelligence 97 (1997) 273--324.

\bibitem{mRMR2}
C.~Ding, H.~Peng, Minimum redundancy feature selection from microarray gene
  expression data, in: Proceedings of the IEEE Computer Society Conference on
  Bioinformatics, CSB'03, IEEE Computer Society, Washington, DC, USA, 2003, pp.
  523--528.

\bibitem{Zhangis_2}
Y.~Zhang, S.~Li, T.~Wang, Z.~Zhang, Divergence-based feature selection for
  separate classes, Neurocomputing 101 (2013) 32--42.

\bibitem{mRMR4}
T.~M. Cover, The best two independent measurements are not the two best, IEEE
  Transactions on Systems, Man, and Cybernetics 4 (1974) 116--117.

\bibitem{mRMR14}
A.~K. Jain, R.~P.~W. Duin, J.~Mao, Statistical pattern recognition: A review,
  IEEE Transactions on Pattern Analysis and Machine Intelligence 22~(1) (2000)
  4--37.

\bibitem{cluster_based}
Q.~Song, J.~Ni, G.~Wang, A fast clustering-based feature subset selection
  algorithm for high-dimensional data, IEEE Transactions on Knowledge and Data
  Engineering 25~(1) (2013) 1--14.

\bibitem{CMIM_WANG}
G.~Wang, F.~H. Lochovsky, Q.~Yang, Feature selection with conditional mutual
  information maximin in text categorization, in: Proceedings of the 19th ACM
  International Conference on Information and Knowledge Management, CIKM'04,
  ACM Press, New York, NY, USA, 2004, pp. 342--349.

\bibitem{mRR}
J.~M. Sotoca, F.~Pla, Supervised feature selection by clustering using
  conditional mutual information-based distances, Pattern Recognition 43~(6)
  (2010) 2068--2081.

\bibitem{Zhangis_1}
Y.~Zhang, Z.~Zhang, Feature subset selection with cumulate conditional mutual
  information minimization, Expert Systems with Applications 39 (2012)
  6078--6088.

\bibitem{JMI}
H.~H. Yang, J.~Moody, Feature selection based on joint mutual information, in:
  Proceedings of International ICSC Symposium on Advances in Intelligent Data
  Analysis, 1999, pp. 22--25.

\bibitem{DISR}
P.~Meyer, G.~Bontempi, On the use of variable complementarity for feature
  selection in cancer classification, Evolutionary Computation and Machine
  Learning in Bioinformatics 3907 (2006) 91--102.

\bibitem{FEAST}
G.~Brown, A.~Pocock, M.-J. Zhao, M.~Luj\'{a}n, Conditional likelihood
  maximisation: A unifying framework for information theoretic feature
  selection, Journal of Machine Learning Research 13 (2012) 27--66.

\bibitem{thirty_years_on}
W.~D. Cooka, L.~M. Seiford, Data envelopment analysis (dea) -- thirty years on,
  European Journal of Operational Research 192 (2009) 1--17.

\bibitem{Ensembles_DEA}
Z.~Zheng, B.~Padmanabhan, Constructing ensembles from data envelopment
  analysis, INFORMS Journal on Computing 19~(4) (2007) 486--496.

\bibitem{DEA_Classifier_1}
H.~Yan, Q.~Wei, Data envelopment analysis classification machine, Information
  Sciences 181 (2011) 5029--5041.

\bibitem{DEA_Inter_Classifier}
P.~C. Pendharkar, M.~D. Troutt, Interactive classification using data
  envelopment analysis, Annals of Operations Research 214 (2014) 125--141.

\bibitem{DEA_Classifier_2}
P.~Pendharkar, Fuzzy classification using the data envelopment analysis,
  Knowledge-Based Systems 31 (2012) 183--192.

\bibitem{CCR}
A.~Charnes, W.~W. Cooper, E.~Rhodes, Measuring the efficiency of decision
  making units, European Journal of Operational Research 2 (1978) 429--444.

\bibitem{IAMBs}
I.~Tsamardinos, C.~F. Aliferis, A.~Statnikov, Algorithms for large scale markov
  blanket discovery, in: Proceedings of the 16th International Florida
  Artificial Intelligence Research Society Conference, FLAIRS'03, AAAI Press,
  Menlo Park, CA, USA, 2003, pp. 376--381.

\bibitem{IAMB2010}
I.~Tsamardinos, C.~Aliferis, A.~Statnikov, Local causal and markov blanket
  induction for causal discovery and feature selection for classification part
  i: algorithms and empirical evaluation, Journal of Machine Learning Research
  11 (2010) 171--234.

\bibitem{karmarkar}
N.~Karmarkar, A new polynomial time algorithm for linear programming,
  Combinatorica 4 (1984) 373--395.

\bibitem{MDL}
U.~M. Fayyad, K.~B. Irani, Multi-interval discretization of continuous valued
  attributes for classification learning, in: Proceedings of the 13th
  International Joint Conference on Artificial Intelligence, IJCAI'93, 1993,
  pp. 1022--1027.

\bibitem{SVM}
N.~Cristianini, J.~Shawe-Taylor, An Introduction to Support Vector Machines and
  Other Kernel-Based Learning Methods, Cambridge University Press, Cambridge,
  UK, 2000.

\bibitem{5}
R.~Quinlan, C4.5: Programs for Machine Learning, Morgan Kaufmann Publishers,
  San Mateo, CA, USA, 1993.

\bibitem{kNN}
D.~Aha, D.~Kibler, Instance-based learning algorithms, Machine Learning 6
  (1991) 37--66.

\bibitem{Weka}
H.~I. Witten, E.~Frank, Data mining: practical machine learning tools and
  techniques with Java implementations, Morgan Kaufmann, San Francisco, CA,
  USA, 2000.

\bibitem{ReliefF}
M.~Robnik-Sikonja, I.~Kononenko, Theoretical and empirical analysis of relief
  and relieff, Machine Learning 53 (2003) 23--69.

\bibitem{rule_of_thumb_DEA}
W.~W. Cooper, L.~M. Seiford, K.~Tone, Data envelopment analysis: A
  comprehensive text with models, applications, references, and DEA-solver
  software, Springer, New York, USA, 2007.

\bibitem{Enhanced_Russell}
J.~T. Pastor, J.~L. Ruiz, I.~Sirvent, An enhanced dea russell graph efficiency
  measure, European Journal of Operational Research 115~(3) (1999) 596--607.

\bibitem{RAM}
W.~W. Cooper, K.~S. Park, J.~T. Pastor, Ram: A range adjusted measure of
  inefficiency for use with additive models, and relations to other models and
  measure in dea, Journal of Productivity Analysis 11 (1999) 5--42.

\bibitem{BAM}
W.~W. Cooper, J.~T. Pastor, F.~Borras, J.~Aparicio, D.~Pastor, Bam: a bounded
  adjusted measure of efficiency for use with bounded additive models, Journal
  of Productivity Analysis 35~(2) (2011) 85--94.

\end{thebibliography}







\end{document}